\documentclass{article}
\usepackage{amsmath}

\PassOptionsToPackage{numbers, compress}{natbib}

\usepackage[eandd, preprint]{neurips_2026}

\usepackage[utf8]{inputenc} 
\usepackage[T1]{fontenc}    
\usepackage[colorlinks=true,linkcolor=black,citecolor=black,urlcolor=black]{hyperref}
\usepackage{url}            
\usepackage{booktabs}       
\usepackage{amsfonts}       
\usepackage{nicefrac}       
\usepackage{microtype}      
\usepackage{xcolor}         
\usepackage{multicol}
\usepackage{longtable}
\usepackage{multirow}
\usepackage{graphicx}
\usepackage{tikz}
\usetikzlibrary{shadows}
\usetikzlibrary{arrows.meta, positioning}

\title{V4FinBench: Benchmarking Tabular Foundation Models, LLMs, and Standard Methods on Corporate Bankruptcy Prediction}

\author{%
  Marcin Kostrzewa$^{1}$ \quad Sebastian Tomczak$^{1}$ \quad Roman Furman$^{3}$ \quad Anna Poberezhna$^{1}$ \\
  \textbf{Michał Furgała}$^{1}$ \quad \textbf{Julia Farganus}$^{1}$ \quad \textbf{Oleksii Furman}$^{1}$ \quad \textbf{Maciej Zięba}$^{1,2}$ \\[2pt]
  $^{1}$Department of Artificial Intelligence, Wrocław University of Science and Technology, Poland \\
  $^{2}$Tooploox, Wrocław, Poland \\
  $^{3}$Opera, Wrocław, Poland \\
  \texttt{marcin.kostrzewa@pwr.edu.pl}
}

\begin{document}

\maketitle

\begin{abstract}
Corporate bankruptcy prediction is a high-stakes financial task characterized by severe class imbalance and multi-horizon forecasting demands. Public datasets supporting it remain scarce and small: widely used free benchmarks contain between 6,000 and 80,000 company-year observations, while larger resources are behind subscription paywalls. To address this gap, we introduce V4FinBench, a benchmark of over one million company-year records from the Visegrád Group (V4) economies (2006--2021), with 131 financial and non-financial features, six prediction horizons, and a composite distress criterion jointly capturing solvency, profitability, and liquidity deterioration. V4FinBench is designed to support the evaluation of tabular and foundation-model methods under realistic class imbalance, with positive rates between 0.19\% and 0.36\%. We provide reference evaluations of standard tabular baselines, finetuned TabPFN, and QLoRA-finetuned Llama-3-8B. With imbalance-aware finetuning, TabPFN matches or exceeds gradient boosting at longer time horizons on both $F_1$-score and ROC-AUC. In contrast, Llama-3-8B trails gradient boosting on ROC-AUC at every horizon and is generally weaker on $F_1$, with the gap widening sharply beyond the immediate horizon. In an external evaluation on the American Bankruptcy Dataset, the V4FinBench-finetuned TabPFN checkpoint improves over vanilla TabPFN, suggesting that adaptation captures transferable financial-distress structure rather than only V4-specific patterns. V4FinBench is publicly released to support further evaluation and development of prediction methods on realistic financial data.
\end{abstract}

\section{Introduction}

Predicting corporate bankruptcy has direct consequences for investors, creditors, and regulators, and is technically demanding: real-world bankruptcy and severe distress events are rare, and useful predictions are required across multiple horizons, from immediate distress detection to several years ahead. Despite its practical importance, the public resources available for studying it remain scarce. The widely used free benchmarks, e.g., the UCI Polish dataset~\cite{tomczak2016polish}, the Taiwanese Bankruptcy Prediction dataset~\cite{liang2016financial}, and the American stock-market dataset~\cite{lombardo2022machine}, contain between 6{,}000 and 80{,}000 company-year observations. Larger and richer resources sit behind subscription paywalls~\cite{shi2024datasets}. This limits reproducibility and constrains the evaluation of modern methods, especially foundation models that require substantially more data for adaptation.

This gap is increasingly important. Tabular foundation models and large language models are becoming plausible candidates for financial prediction, but existing public bankruptcy datasets are generally too small to finetune them meaningfully and too limited in horizon coverage to expose horizon-dependent behavior. Evaluating such models under realistic financial conditions therefore requires a benchmark that combines scale, severe imbalance, multi-horizon labels, and a reproducible evaluation protocol.

We introduce \textbf{V4FinBench}, a public benchmark of 1{,}106{,}879 company-year observations from the Visegrád Group (V4) economies: Czech Republic, Hungary, Poland, and Slovakia. The benchmark spans 2006--2021, includes 203{,}900 unique companies and 131 financial and non-financial features, and provides six prediction-horizon tasks from current-year distress detection to five-year-ahead prediction. Labels are based on a composite financial distress criterion requiring simultaneous deterioration in solvency, profitability, and liquidity. The resulting tasks exhibit the severe class imbalance characteristic of the underlying problem, with positive rates between 0.19\% and 0.36\%. V4FinBench is publicly released on Kaggle with documentation, and the label-construction code is released in the accompanying repository\footnote{\url{https://www.kaggle.com/datasets/sebastiantomczak10/v4-group-corporate-bankruptcy/data}; \url{https://github.com/genwro-ai/V4FinBench}}.

Beyond releasing the data, we define a fixed evaluation protocol with company-level grouped, country-stratified folds, validation-based threshold calibration, and standard metrics for imbalanced prediction. This protocol is intended to make V4FinBench a reusable benchmark for comparing classical tabular methods, tabular foundation models, and LLM-based approaches under the same conditions.

Using V4FinBench, we conduct reference evaluations of standard baselines and two foundation-model approaches: TabPFN finetuned with imbalance-aware context construction and Llama-3-8B finetuned with QLoRA on serialized financial records. Three findings emerge. First, with prototype-undersampling context construction, TabPFN matches or exceeds gradient-boosted trees at longer horizons on both $F_1$-score and ROC-AUC. Second, QLoRA-finetuned Llama-3-8B trails gradient boosting on ROC-AUC at every horizon and is generally weaker on $F_1$, with the gap widening beyond the immediate horizon. Third, V4FinBench-finetuned TabPFN improves over vanilla TabPFN on the American Bankruptcy Dataset, suggesting that the benchmark can support adaptation that transfers beyond the original V4 distribution.

We make three contributions:
\begin{itemize}
\item \textbf{V4FinBench}, a large public corporate-distress benchmark designed to support foundation-model finetuning, covering 1.1M company-year observations across four countries, six prediction horizons, and a composite distress criterion.
\item \textbf{A reproducible evaluation protocol and reference results} for classical baselines, TabPFN with imbalance-aware context construction, and QLoRA-finetuned Llama-3-8B, providing a point of comparison for future methods.
\item \textbf{A transfer analysis} evaluating whether V4FinBench-finetuned TabPFN improves performance on an external financial-distress dataset, probing whether adaptation captures reusable financial structure rather than only V4-specific patterns.
\end{itemize}

\section{Related work}

\subsection{Public datasets and benchmarks for bankruptcy prediction}

A recent taxonomy by~\cite{shi2024datasets} surveys the landscape of public bankruptcy-prediction resources and highlights a shortage of large, freely accessible datasets. The most widely used public benchmarks are the UCI Polish Companies Bankruptcy dataset~\cite{tomczak2016polish}, which provides roughly 10{,}000 company-year records across five forecasting horizons, the Taiwanese Bankruptcy Prediction dataset~\cite{liang2016financial}, with about 6{,}800 company-year observations and a single-period label derived from Taiwan Stock Exchange regulations, and the American stock-market dataset~\cite{lombardo2022machine}, comprising approximately 78{,}000 company-year records of US-listed companies, with bankruptcy defined via Chapter 7 or Chapter 11 filings. 

Larger or more feature-rich resources, such as Compustat, Bureau van Dijk's Orbis and Amadeus, and EMIS, exist but are subscription-based and therefore restrict reproducibility~\cite{shi2024datasets}. Studies in specific regions, including the Visegr\'{a}d Group \cite{tomczak2025identification}, typically operate on country-specific samples of a few thousand firms with heterogeneous distress definitions~\cite{Durica_Frnda_Svabova_2023,Gavurova_Jencova_Bacik_Miskufova_Letkovsky_2022,tomczak2025,Valaskova_Gajdosikova_Belas_2023}, well below the scale needed for foundation-model finetuning. V4FinBench addresses this gap by providing a large public benchmark with a multi-horizon construction and a reproducible evaluation protocol.

\subsection{Methods for bankruptcy prediction}

Statistical bankruptcy prediction has a long history beginning with discriminant-analysis models such as Altman's Z-score~\cite{altman1968financial} and Ohlson's O-score~\cite{ohlson1980financial}. The field has progressively shifted toward machine-learning approaches, including random forests, support vector machines, and gradient boosting~\cite{barboza2017machine}, and more recently toward deep models leveraging textual disclosures and earnings calls~\cite{mai2019deep,matin2019predicting}. Across the modern literature, gradient-boosted trees (XGBoost, CatBoost, LightGBM) are consistently among the strongest performers on tabular financial features and serve as standard reference baselines~\cite{Alanis2023bench,barboza2017machine}. For this reason, we treat them as the primary reference methods in V4FinBench, alongside logistic regression, random forest, and multilayer perceptron.

\subsection{Foundation models for structured financial prediction}

Foundation-model approaches in finance have mostly focused on text, including BloombergGPT~\cite{wu2023bloomberggpt}, FinBERT~\cite{araci2019finbert,yang2020finbert}, FinGPT~\cite{yang2023fingpt}, PIXIU~\cite{xie2023pixiu}, and InvestLM~\cite{yang2023investlm}. Structured prediction has been explored through tabular foundation models such as TabPFN~\cite{hollmann2025tabpfn}, which performs in-context prediction over tabular examples, and through LLMs applied to serialized tabular rows, as in TabLLM~\cite{hegselmann2023tabllm}. Our prior work~\cite{kostrzewa2025foundation} evaluated API-served Llama-3.3-70B and pretrained TabPFN on a limited subsample of this data. V4FinBench releases the underlying dataset publicly for the first time, and the present work extends the evaluation with a fixed multi-horizon benchmark protocol and finetuning recipes for both TabPFN and Llama-3, showing that imbalance-aware TabPFN adaptation changes the comparison with gradient-boosted baselines at longer horizons.
 
\section{V4FinBench}
\label{sec:dataset}

\begin{figure}[t]
\centering
\begin{tikzpicture}[
    node distance=0.9cm,
    every node/.style={font=\small},
    source/.style={
        draw=blue!50,
        rounded corners=4pt,
        align=center,
        minimum width=3.4cm,
        minimum height=0.95cm,
        fill=blue!8,
        line width=0.6pt,
        drop shadow={shadow xshift=0.6pt, shadow yshift=-0.6pt, opacity=0.15}
    },
    process/.style={
        draw=teal!55,
        rounded corners=4pt,
        align=center,
        minimum width=3.4cm,
        minimum height=0.95cm,
        fill=teal!8,
        line width=0.6pt,
        drop shadow={shadow xshift=0.6pt, shadow yshift=-0.6pt, opacity=0.15}
    },
    output/.style={
        draw=orange!60,
        rounded corners=4pt,
        align=center,
        minimum width=3.4cm,
        minimum height=0.95cm,
        fill=orange!8,
        line width=0.6pt,
        drop shadow={shadow xshift=0.6pt, shadow yshift=-0.6pt, opacity=0.15}
    },
    arrow/.style={
        ->,
        >=stealth,
        line width=0.7pt,
        draw=gray!75
    },
    grouplabel/.style={
        font=\footnotesize\itshape,
        text=gray!70
    }
]
\node[source] (raw) {EMIS financial\\statements};
\node[process, right=0.9cm of raw] (items) {Raw accounting items\\\& company metadata};
\node[output, right=0.9cm of items] (features) {131 financial \&\\non-financial features};

\node[process, below=1.0cm of items] (label) {Composite distress label\\\textit{\footnotesize solvency $\cdot$ profitability $\cdot$ liquidity}};
\node[output, below=1.0cm of raw] (tasks) {Six horizon tasks\\$h \in \{0, \ldots, 5\}$};
\node[output, below=1.0cm of features] (eval) {Released folds \&\\reference evaluations};

\draw[arrow] (raw) -- (items);
\draw[arrow] (items) -- (features);
\draw[arrow] (items) -- (label);
\draw[arrow] (label) -- (tasks);
\draw[arrow] (label) -- (eval);
\draw[arrow] (features) -- (eval);

\node[grouplabel, left=0.3cm of raw, anchor=east] {\textsc{source}};
\node[grouplabel, left=0.3cm of tasks, anchor=east] {\textsc{output}};

\end{tikzpicture}
\caption{V4FinBench overview. Raw EMIS financial statements and company metadata are transformed into 131 features, labeled using a composite solvency-profitability-liquidity distress criterion, and released as six horizon-specific benchmark tasks with fixed folds and reference evaluations.}
\label{fig:v4finbench-overview}
\end{figure}
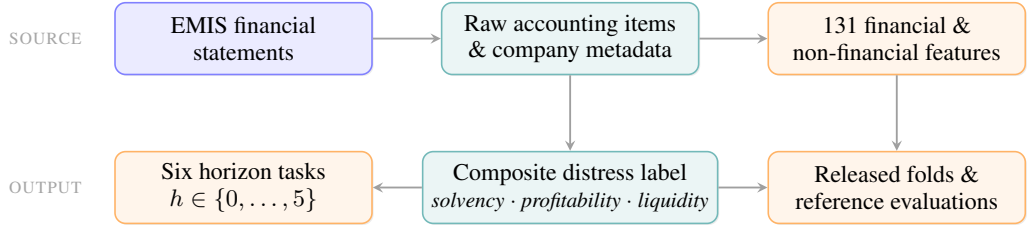

This section describes V4FinBench: the underlying data source and coverage, the composite distress definition used to construct labels, the procedure for generating six prediction-horizon tasks, and the intended use of the benchmark. Figure~\ref{fig:v4finbench-overview} summarizes the benchmark construction pipeline and released evaluation setup.

\subsection{Data source and coverage}
\label{sec:data-source}

V4FinBench is constructed from company-level financial statement data extracted from the EMIS service (\url{https://www.emis.com/}), together with company metadata such as country, legal form, industry classification, operational status, and number of employees. V4FinBench covers companies registered in Poland, Hungary, Czech Republic, and Slovakia over 2006--2021. It contains 203{,}900 unique companies and 1{,}106{,}879 company-year observations, each described by 131 financial and non-financial features. These features cover company metadata, financial ratios, growth measures, size measures, and sector-relative indicators. The complete feature list, country-level counts, and data-consistency checks are provided in Appendix~\ref{app:data-source}.

\subsection{Distress definition}
\label{sec:distress-def}

V4FinBench labels are derived from a composite financial distress criterion applied to a company's final available annual report. A company is labeled distressed in its final reporting year if it simultaneously satisfies all three of the following conditions:

\begin{itemize}
    \item \textbf{Solvency:} negative equity to total assets ($\textit{equity}/\textit{total\_assets} < 0$),
    \item \textbf{Profitability:} negative EBITDA to total assets ($\textit{EBITDA}/\textit{total\_assets} < 0$),
    \item \textbf{Liquidity:} current ratio below 0.6 ($\textit{current\_assets}/\textit{current\_liabilities} < 0.6$).
\end{itemize}

The criterion is designed to identify companies in severe financial distress rather than firms that are weak in only one dimension. Companies whose final available report is from 2021 are excluded from the dataset since the dataset terminates in 2021, and we cannot verify whether such a company genuinely stopped reporting or simply continued reporting beyond our window. All remaining companies that do not
satisfy the criterion are labeled non-distressed. Appendix~\ref{app:distress-definition} describes how this composite definition relates to alternative criteria used in prior V4-region studies.

\subsection{Multi-horizon construction}
\label{sec:multi-horizon}

V4FinBench provides six binary prediction tasks corresponding to horizons $h \in \{0,1,\ldots,5\}$ years ahead. For each horizon $h$, companies meeting the composite distress criterion have their final $h$ years of data removed, and the resulting final observation receives a positive label. All other company-year observations are assigned a negative label.

\begin{table}[ht]
\centering
\caption{Overview of datasets and prediction tasks.}
\label{tab:datasets}
\begin{tabular}{l r r r}
\toprule
\textbf{Prediction Horizon} & \textbf{Total Instances} & \textbf{Distressed} & \textbf{Non-distressed}  \\
\midrule
 0 years (current)   & 1,000,087 & 3,587 & 996,500 \\
 1 year ahead        &   996,500 & 3,054 & 993,446  \\
 2 years ahead       &   898,692 & 2,374 & 896,318 \\
 3 years ahead       &   793,234 & 1,896 & 791,338 \\
 4 years ahead       &   700,041 & 1,485 & 698,556 \\
 5 years ahead       &   598,832 & 1,154 & 597,678 \\
\bottomrule
\end{tabular}
\end{table}

As shown in Table~\ref{tab:datasets}, both total size and the number of positives decrease with horizon length, reflecting the need for sufficient reporting history before the distress event. Across horizons, the positive-class rate ranges from 0.19\% to 0.36\%. The release includes the base company-year records, six labeled task files, fold assignments, and label-construction code.

\paragraph{Intended use.}
V4FinBench is intended for evaluating machine-learning methods for structured corporate financial-distress prediction under realistic class imbalance and multi-horizon forecasting. It supports comparisons under the released folds, preprocessing protocol, and composite distress definition. It is not intended to support claims about formal legal bankruptcy filings, individual credit decisions, or deployment-ready risk scoring without additional jurisdiction-specific validation, fairness analysis, and human oversight.

\section{Evaluation protocol}
\label{sec:eval-protocol}

We define a fixed protocol for reproducible evaluation. We use 5-fold stratified cross-validation with company-level grouping within country: all observations from a company are assigned to the same fold, and country proportions are preserved across folds. In each iteration, one fold is used for testing, one for validation, and the remaining three for training, yielding an approximate 60/20/20 split. Fold indices are released and shared across all methods.

Missing values are imputed using training-set medians, and features are standardized using training-set statistics. Both transformations are computed separately within each fold to prevent leakage. We report accuracy, precision, recall, $F_1$, and ROC-AUC, averaged across folds with standard deviations. Given the 0.19--0.36\% positive rate, $F_1$ and ROC-AUC are the primary metrics.

\section{Methods}
\label{sec:methods}
 
We evaluate three families of methods on V4FinBench: tabular foundation
models (TabPFN with imbalance-aware in-context finetuning), large language
models (Llama-3-8B with QLoRA on serialized financial records), and
classical tabular baselines.
 
\subsection{TabPFN with imbalance-aware finetuning}
\label{sec:tabpfn}
 
TabPFN~\cite{hollmann2025tabpfn} operates as an in-context learner: at inference time, it conditions predictions for a query set on a context set of labeled examples passed directly into the model. The context is bounded
-- in our experiments to 10{,}000 samples -- which means the composition of the context determines what signal the model sees. Under V4FinBench's class distribution, a randomly drawn context of 10{,}000 samples contains only tens of positive examples on average, leaving the minority class weakly represented relative to the majority. Context composition is therefore a crucial choice.

\begin{figure}
\centering
\begin{tikzpicture}[scale=0.9, transform shape,
    pool/.style={draw=gray!50, line width=0.3pt, rounded corners=4pt, fill=gray!3},
    proto_box/.style={draw=teal!60!black, line width=0.3pt, rounded corners=4pt, fill=teal!3},
    flow/.style={->, >=latex, gray!60, line width=0.4pt},
    maj/.style={fill=gray!65, draw=none},
    min/.style={fill=orange!75!red, draw=none},
    proto/.style={fill=teal!55!black, draw=none},
    cluster/.style={draw=teal!50!black, line width=0.25pt, dashed, fill=teal!8, fill opacity=0.5},
    title/.style={font=\footnotesize\sffamily, text=black},
    sublabel/.style={font=\scriptsize\sffamily, text=gray!50!black},
    minlabel/.style={font=\scriptsize\sffamily, text=orange!60!red},
]

\node[title] at (7.0, 5.35) {Training pool};

\draw[pool] (4.5, 3.9) rectangle (9.5, 5.1);

\foreach \x in {0,...,16} {
    \foreach \y in {0,...,4} {
        \pgfmathsetmacro{\xpos}{4.72 + \x * 0.22}
        \pgfmathsetmacro{\ypos}{4.05 + \y * 0.22 + (mod(\x,2)*0.03)}
        \fill[maj] (\xpos, \ypos) circle (0.045);
    }
}

\foreach \pos in {(9.05,4.60), (9.15,4.50), (9.22,4.60), (9.10,4.40), (9.20,4.45)} {
    \fill[min] \pos circle (0.06);
}

\draw[gray!50, line width=0.25pt] (4.72, 4.50) -- (3.9, 4.50);
\node[sublabel, anchor=east] at (3.85, 4.50) {Majority};
\draw[orange!50!red, line width=0.25pt] (9.27, 4.50) -- (10.1, 4.50);
\node[minlabel, anchor=west] at (10.15, 4.50) {Minority};

\draw[flow] (5.5, 3.9) -- (2.0, 3.25);
\draw[flow] (7.0, 3.9) -- (7.0, 3.25);
\draw[flow] (8.5, 3.9) -- (12.0, 3.25);

\node[title] at (2.0, 3.45) {No resampling};
\node[title] at (7.0, 3.45) {Random undersampling};
\node[title] at (12.0, 3.45) {Prototype undersampling};

\draw[pool] (-0.4, 1.2) rectangle (4.4, 3.1);

\foreach \x in {0,...,15} {
    \foreach \y in {0,...,7} {
        \pgfmathsetmacro{\xpos}{-0.18 + \x * 0.22}
        \pgfmathsetmacro{\ypos}{1.35 + \y * 0.22 + (mod(\x,2)*0.03)}
        \fill[maj] (\xpos, \ypos) circle (0.045);
    }
}

\foreach \pos in {(3.95,2.20), (4.05,2.10), (4.12,2.20), (4.00,2.00), (4.10,2.05)} {
    \fill[min] \pos circle (0.06);
}

\draw[pool] (4.6, 1.2) rectangle (9.4, 3.1);
\foreach \pos in {(4.85,2.88), (5.25,2.62), (5.65,2.92), (6.05,2.50), (6.45,2.85),
                  (6.85,2.55), (7.15,2.88),
                  (4.95,2.30), (5.40,2.05), (5.80,2.35), (6.20,2.02), (6.60,2.30),
                  (6.95,2.05), (7.20,2.32),
                  (5.05,1.72), (5.50,1.50), (5.90,1.75), (6.30,1.48),
                  (6.70,1.72), (7.05,1.50),
                  (4.88,1.32), (5.55,1.32), (6.20,1.32), (6.85,1.32)} {
    \fill[maj] \pos circle (0.06);
}
\foreach \pos in {(7.80,2.85), (8.10,2.65), (8.40,2.92), (8.72,2.60), (9.02,2.85),
                  (7.90,2.20), (8.22,1.98), (8.55,2.20), (8.88,1.95),
                  (7.95,1.55), (8.27,1.35), (8.62,1.55), (8.95,1.32)} {
    \fill[min] \pos circle (0.06);
}

\draw[proto_box] (9.6, 1.2) rectangle (14.4, 3.1);

\draw[cluster] (10.25, 2.70) ellipse (0.32 and 0.20);
\fill[proto] (10.25, 2.70) circle (0.08);

\draw[cluster] (11.15, 2.70) ellipse (0.32 and 0.20);
\fill[proto] (11.15, 2.70) circle (0.08);

\draw[cluster] (10.70, 2.15) ellipse (0.32 and 0.20);
\fill[proto] (10.70, 2.15) circle (0.08);

\draw[cluster] (11.60, 2.15) ellipse (0.32 and 0.20);
\fill[proto] (11.60, 2.15) circle (0.08);

\draw[cluster] (10.25, 1.60) ellipse (0.32 and 0.20);
\fill[proto] (10.25, 1.60) circle (0.08);

\draw[cluster] (11.15, 1.60) ellipse (0.32 and 0.20);
\fill[proto] (11.15, 1.60) circle (0.08);

\foreach \pos in {(12.85,2.85), (13.15,2.65), (13.45,2.92), (13.72,2.60), (14.00,2.85),
                  (12.95,2.20), (13.25,1.98), (13.58,2.20), (13.88,1.95),
                  (13.00,1.55), (13.30,1.35), (13.62,1.55), (13.95,1.32)} {
    \fill[min] \pos circle (0.06);
}

\node[sublabel] at (2.0, 0.90) {tens of positives in 10,000};
\node[sublabel] at (7.0, 0.90) {$N_{\mathrm{min}}/N_{\mathrm{maj}}=0.3$};
\node[sublabel] at (12.0, 0.90) {$N_{\mathrm{min}}/N_{\mathrm{maj}}=0.3$};

\node[sublabel, text=orange!55!red] at (2.0, 0.62) {Minority swamped};
\node[sublabel] at (7.0, 0.62) {Structure not preserved};
\node[sublabel] at (12.0, 0.62) {Structure preserved};

\fill[maj] (4.6, 0.10) circle (0.06);
\node[sublabel, anchor=west] at (4.75, 0.10) {Majority};

\draw[cluster] (7.0, 0.10) ellipse (0.16 and 0.10);
\fill[proto] (7.0, 0.10) circle (0.07);
\node[sublabel, anchor=west] at (7.25, 0.10) {Prototype};

\fill[min] (9.4, 0.10) circle (0.06);
\node[sublabel, anchor=west] at (9.55, 0.10) {Minority};

\end{tikzpicture}

\caption{TabPFN context construction under severe class imbalance. \textbf{No resampling} preserves the original imbalance, leaving only tens of positives in a 10{,}000-sample context. \textbf{Random undersampling} subsamples majority points arbitrarily until $N_{\mathrm{min}}/N_{\mathrm{maj}}=0.3$. \textbf{Prototype undersampling} uses the same class budget but selects majority representatives via K-means, preserving more non-distressed population structure.}

\label{fig:context-strategies}

\end{figure}
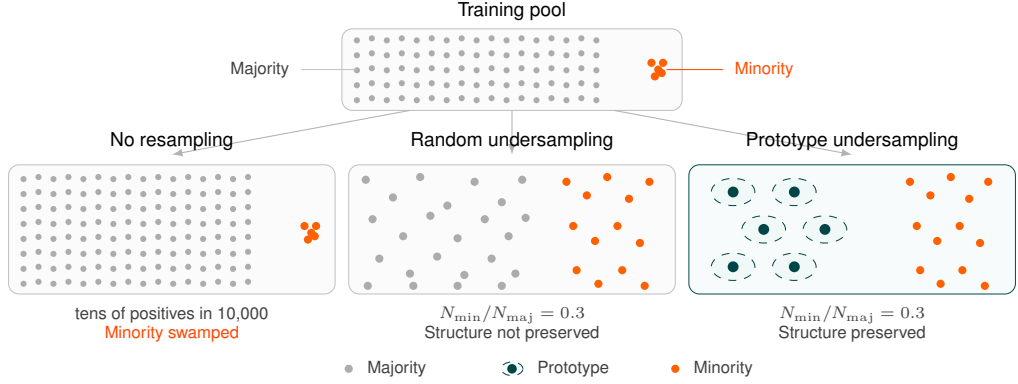

We initialize from the pretrained TabPFN v2 checkpoint and finetune with the Adam optimizer for 10 epochs (full hyperparameters in
Appendix~\ref{app:tabpfn}). Finetuning is performed on a single NVIDIA A100 GPU. At each epoch, a training subset is sampled to form the context and query batches. To address the imbalance problem in context construction, we compare three strategies (Figure~\ref{fig:context-strategies}):

\begin{itemize}
    \item \textbf{No resampling}: the context is sampled uniformly from the training set. Under severe imbalance, the minority class is represented by only tens of examples in a 10{,}000-sample context.

    \item \textbf{Random undersampling}: the majority class is randomly subsampled so that the post-resampling minority-to-majority ratio is $0.3$, with the minority class left untouched. Given V4FinBench's 0.19--0.36\% positive rate, this retains most real positives and pairs them with a majority subset about $3.3\times$ their size. The minority is therefore well represented, but the retained majority samples are drawn arbitrarily and may not span the diversity of the non-distressed population.
    
    \item \textbf{Prototype undersampling}: the same minority-to-majority ratio of $0.3$ is used, but the majority subset is selected by clustering majority-class samples with MiniBatchKMeans and retaining, for each cluster, the real observation closest to the centroid. This yields a compact but representative set of majority-class examples, preserving the structure of the non-distressed population.
\end{itemize}

A checkpoint is saved after each epoch. For each checkpoint, the decision threshold is calibrated on the validation fold by maximizing $F_1$ on the
precision-recall curve, and the calibrated validation $F_1$ is recorded. The checkpoint with the best calibrated validation $F_1$ is retained for test-time evaluation, ensuring that both the model state and the decision threshold are selected jointly under the same criterion used for reporting. The best prototype-undersampling checkpoints for each horizon task are
available on Hugging Face\footnote{\url{https://huggingface.co/Manik2000/v4finbench-tabpfn}}.

\subsection{Llama-3-8B with QLoRA}
\label{sec:llama}

We finetune \texttt{meta-llama/Meta-Llama-3-8B}~\cite{llama3_modelcard} using QLoRA~\cite{dettmers2023qlora}. The base model is loaded in 4-bit precision, and only LoRA adapter matrices are updated. Each company-year record is serialized as an instruction-style conversation. The system prompt asks the model to predict the binary label for the specified prediction horizon and answer with a single word, \texttt{YES} or \texttt{NO}. The user turn contains grouped financial features as labeled key-value pairs, and the assistant turn contains the ground-truth label. The prompt is parameterized by horizon, with the full template provided in Appendix~\ref{app:llama}. Training is performed on a single NVIDIA A100 GPU. Full formatting and training details are provided in Appendix~\ref{app:llama}.

\subsection{Standard baselines}
\label{sec:baselines}
 
We compare against six standard tabular methods: logistic regression (LR),
multilayer perceptron (MLP), random forest, XGBoost~\cite{chen2016xgboost},
CatBoost~\cite{prokhorenkova2018catboost}, and
LightGBM~\cite{ke2017lightgbm}. Hyperparameters for each method are
selected via grid search on the validation fold; full search grids are
provided in Appendix~\ref{app:baselines}.

\section{Experiments}
\label{sec:experiments}

We use V4FinBench to evaluate three questions: whether imbalance-aware TabPFN finetuning can compete with strong classical baselines, whether an instruction-finetuned LLM can solve the same structured prediction task from serialized records, and whether V4FinBench finetuning transfers beyond the original V4 distribution.

\paragraph{Shared setup.}
Unless stated otherwise, experiments follow the evaluation protocol in Section~\ref{sec:eval-protocol}. For each method and fold, hyperparameters are selected on the validation fold. Because the tasks are severely imbalanced, the classification threshold is also selected on the validation fold by maximizing $F_1$ on the precision-recall curve, then applied unchanged to the test fold. We report fold-averaged metrics with standard deviations. The Llama-3-8B experiment deviates from the full protocol due to compute constraints, as described in Section~\ref{sec:exp-llama}.
 
\subsection{Standard methods vs.\ finetuned TabPFN}
\label{sec:exp-tabpfn}

\begin{figure}[ht]
    \centering
    \includegraphics[width=0.85\linewidth]{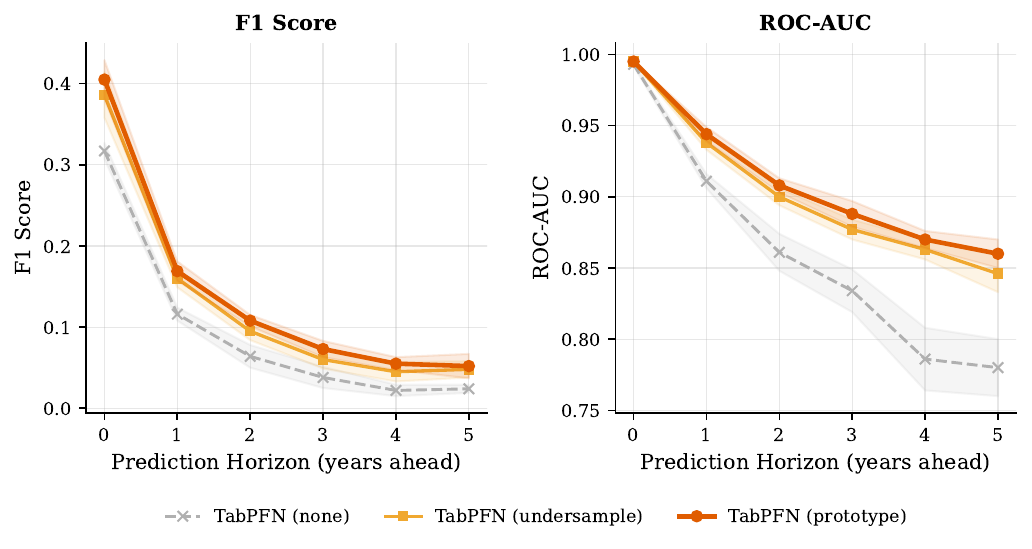}
    \caption{TabPFN context-construction ablation across prediction horizons. Prototype undersampling performs best or near-best across horizons, followed by random undersampling and no resampling.}
    \label{fig:tabpfn-ablation}
\end{figure}

\begin{figure}[ht]
 \centering
\includegraphics[width=0.85\linewidth]{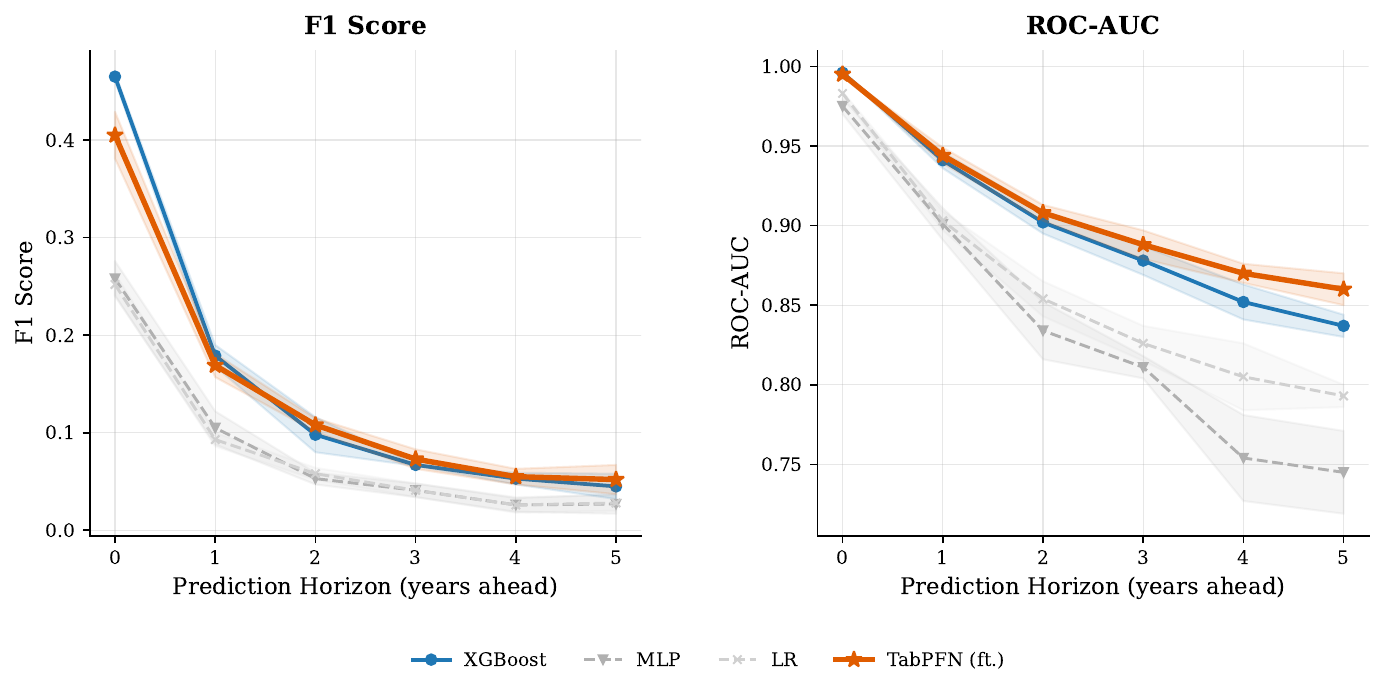}
\caption{Finetuned TabPFN (prototype undersampling) against XGBoost and representative standard baselines. Lines show fold-averaged metrics;
shaded bands show $\pm 1$ standard deviation across folds.}
\label{fig:tabpfn-vs-baselines}
\end{figure}

We compare the three TabPFN context-construction variants from Section~\ref{sec:tabpfn} with the six classical baselines from Section~\ref{sec:baselines} on all six prediction horizons. This experiment evaluates whether TabPFN can be adapted to a large, severely imbalanced financial benchmark and whether context construction affects its competitiveness. Figure~\ref{fig:tabpfn-ablation} reports the context-construction ablation, while Figure~\ref{fig:tabpfn-vs-baselines} compares the best TabPFN variant with the strongest classical baselines. Full per-horizon results are reported in Appendix~\ref{app:full-results}.
 
\paragraph{Findings.}
Figure~\ref{fig:tabpfn-ablation} compares the three TabPFN context-construction strategies. Prototype undersampling gives the best results on both $F_1$ and ROC-AUC, followed by random undersampling and then no resampling. Figure~\ref{fig:tabpfn-vs-baselines} compares prototype-undersampled TabPFN with the strongest classical baselines. Gradient-boosted trees form a strong baseline cluster and consistently outperform logistic regression, random forests, and MLPs. With prototype undersampling, TabPFN closes most of the $F_1$ gap at $h=0$ and surpasses the strongest gradient-boosted baseline from $h=2$ onward. On ROC-AUC, prototype-undersampled TabPFN matches or exceeds gradient boosting at every horizon. The gap between prototype and random undersampling indicates that preserving majority-class structure matters beyond simply increasing minority exposure.

\subsection{QLoRA-finetuned Llama-3-8B vs. XGBoost}
\label{sec:exp-llama}
 \paragraph{Setup.}
Compute constraints prevent following the full protocol of Section~\ref{sec:eval-protocol} for Llama-3-8B. We therefore deviate as follows. \emph{(i) Training data}: for each horizon, a stratified subset of 20,000 observations is split 80/20 into training and validation, rather than using 5-fold cross-validation over the full benchmark. \emph{(ii) Test data}: for each horizon, the test set contains all available positive held-out instances and a sampled subset of negative held-out instances, up to 100{,}000 total observations. \emph{(iii) Prediction extraction}: we extract token-level log-probabilities for the \texttt{YES} and \texttt{NO} tokens to obtain soft probability estimates. The decision threshold is selected on the validation split by maximizing $F_1$ and then applied to the test set. To enable a like-for-like comparison, we train a dedicated XGBoost classifier on the identical training, validation, and test rows used for Llama-3-8B, with median imputation. 
 
\paragraph{Findings.}
Table~\ref{tab:llama-vs-xgb} reports the probability-based comparison using token-level \texttt{YES}/\texttt{NO} log-probabilities and validation-calibrated thresholds. XGBoost outperforms QLoRA-finetuned Llama-3-8B at every horizon on ROC-AUC and generally on $F_1$. The Llama model performs best at the immediate horizon, but its ranking ability declines sharply after $h=0$ and remains close to chance at longer horizons. Overall, probability-based evaluation recovers some signal from Llama-3-8B, but the model remains far below XGBoost on this structured prediction task.
 
\begin{table}[t]
\centering
\caption{QLoRA-finetuned Llama-3-8B (probability-based) vs.\ XGBoost on identical train/test rows.}
\label{tab:llama-vs-xgb}
\small
\begin{tabular}{ccccc}
\toprule
& \multicolumn{2}{c}{\textbf{ROC-AUC}} & \multicolumn{2}{c}{\textbf{$F_1$}} \\
\cmidrule(lr){2-3} \cmidrule(lr){4-5}
\textbf{Horizon} & Llama-3-8B & XGBoost & Llama-3-8B & XGBoost \\
\midrule
0 & 0.825 & 0.995 & 0.308 & 0.483 \\
1 & 0.568 & 0.937 & 0.095 & 0.218 \\
2 & 0.597 & 0.908 & 0.119 & 0.113 \\
3 & 0.517 & 0.879 & 0.042 & 0.055 \\
4 & 0.583 & 0.857 & 0.011 & 0.040 \\
5 & 0.553 & 0.811 & 0.030 & 0.037 \\
\bottomrule
\end{tabular}
\end{table}

\subsection{Cross-dataset transfer to the American Bankruptcy Dataset}
\label{sec:exp-transfer}
 
In this subsection, we check whether the finetuning adaptation captures the transferable structure of the underlying domain or merely fits the idiosyncrasies of the training distribution. We probe this by evaluating a V4FinBench-finetuned TabPFN checkpoint on a separate public benchmark drawn from a different economy, period, and feature construction.
 
\paragraph{Setup.}
We use the American Bankruptcy Dataset~\cite{lombardo2022machine}, which
contains $\sim$78{,}000 firm-year observations from US-listed companies. We compare two configurations of TabPFN: \emph{vanilla}, a non-finetuned TabPFN, and \emph{V4FinBench-finetuned} (the prototype-undersampling variant from
Section~\ref{sec:exp-tabpfn}), taken at the one-year prediction horizon ($h{=}1$),
which is the most directly comparable setting to the American dataset's
one-year-ahead label. Threshold calibration is performed on the American
validation split, so no V4FinBench-specific thresholds transfer across datasets.
 
\begin{figure}[ht]
    \centering
    \includegraphics[height=7cm]{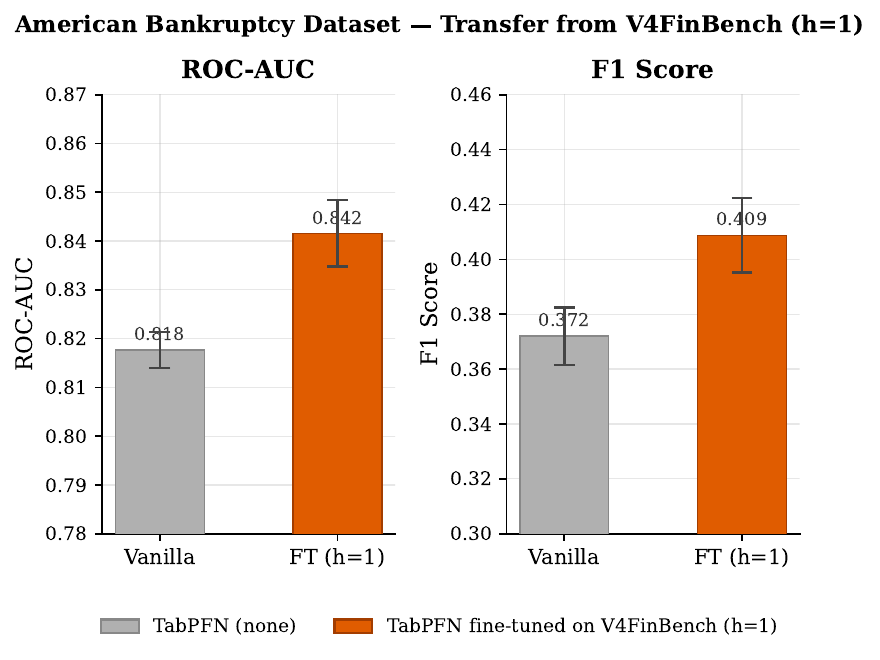}
\caption{Transfer of V4FinBench-finetuned TabPFN to the American
Bankruptcy Dataset~\cite{lombardo2022machine}. \emph{Vanilla} denotes
the pretrained TabPFN without V4FinBench finetuning. \emph{FT (h$=$1)}
denotes the prototype-undersampled checkpoint trained on the
V4FinBench one-year-ahead task. Finetuning improves both ROC-AUC
($0.818 \rightarrow 0.842$) and $F_1$ ($0.372 \rightarrow 0.409$).
Error bars denote variability across runs.}
    \label{fig:american-transfer}
\end{figure}

\paragraph{Findings.}
Figure~\ref{fig:american-transfer} reports ROC-AUC and $F_1$ on the
American test set. The V4FinBench-finetuned ($h{=}1$) checkpoint
outperforms vanilla TabPFN on both metrics: ROC-AUC improves from
$0.818$ to $0.842$ and $F_1$ improves from $0.372$ to $0.409$. Because
the American dataset differs in economy, feature construction, and distress definition, this result suggests that V4FinBench finetuning captures transferable financial-distress structure rather than only V4-specific patterns.
 
\section{Limitations}
\label{sec:limitations}

V4FinBench is designed as a public benchmark for corporate distress prediction, but its scope and experimental results have several limitations.

\paragraph{Geographic scope.}
V4FinBench covers four Central European economies: Poland, Hungary, the Czech Republic, and Slovakia. Although these countries provide a diverse regional setting, models trained or selected on V4FinBench may not transfer to economies with substantially different accounting standards, reporting practices, ownership structures, or insolvency regimes. Our transfer experiment on the American Bankruptcy Dataset provides one external validation point, but broader geographic robustness remains an open question.

\paragraph{Composite distress labels.}
The benchmark labels financial distress rather than formal legal bankruptcy filings. A positive label requires simultaneous deterioration in solvency, profitability, and liquidity, which provides a uniform financial-statement-based criterion across countries. The trade-off is that V4FinBench evaluates the prediction of a financial-state condition, not a legal bankruptcy event. Users interested in filing-based prediction should treat the released labels as one possible operational definition and may construct alternative labels from the released data.

\paragraph{Responsible use.}
V4FinBench is intended for benchmarking and research, not for direct automated decision-making. Bankruptcy-risk and distress-prediction models can affect access to credit, supplier relationships, investment decisions, or regulatory attention. Any deployment would require additional validation, monitoring, fairness analysis across sectors and countries, and human oversight.

\section{Conclusions}
\label{sec:conclusions}

We introduced V4FinBench, a public benchmark of 1.1M company-year records from the Visegrád Group economies, with 131 features, six prediction horizons, released folds, and a composite solvency-profitability-liquidity distress criterion. The benchmark is designed to support reproducible evaluation of corporate distress prediction methods under realistic class imbalance and multi-horizon forecasting.

Our reference evaluations show three main findings. First, prototype-undersampled TabPFN matches or exceeds gradient-boosted trees on ROC-AUC at every horizon and on $F_1$ from $h{=}2$ onward, showing that context construction is important when applying tabular foundation models to severely imbalanced financial data. Second, in the probability-based diagnostic comparison, QLoRA-finetuned Llama-3-8B trails the matched XGBoost baseline on ROC-AUC at every horizon and is generally weaker on $F_1$, suggesting that serialized-record LLM finetuning is not yet competitive with strong tabular baselines in this setup. Third, the V4FinBench-finetuned one-year-ahead TabPFN checkpoint improves upon the vanilla TabPFN on the American Bankruptcy Dataset, providing preliminary evidence that the benchmark can support adaptation beyond the original V4 distribution.
 
V4FinBench is intended to open a reproducible path for future work on corporate financial-distress prediction at a realistic scale. By releasing the data, horizon-specific tasks, folds, and reference results, the benchmark enables direct comparison of new methods under severe class imbalance and multi-horizon forecasting. Natural extensions include testing additional tabular foundation models, larger or financially specialized LLMs, multimodal approaches that combine structured financial indicators with textual disclosures, alternative distress definitions, and longer-horizon studies as additional reporting years become available.

\section*{Acknowledgements}
The research in this paper has been partially supported by the National Science Centre (NCN, Poland), under Grant no. 2020/39/D/HS4/02384 and  under Grant no. 2024/55/B/ST6/02100.

We would like to also gratefully acknowledge the Polish high-performance computing infrastructure PLGrid (HPC Center: ACK Cyfronet AGH) for providing computer facilities and support within computational grants no. PLG/2025/018494 and no. PLG/2026/019281.

\bibliographystyle{plainnat}
\bibliography{bib}

\appendix

\section{TabPFN tuning configuration}
\label{app:tabpfn}

Table~\ref{tab:tabpfn_config} shows the configuration used for TabPFN finetuning across all tasks.

\begin{table}[ht]
    \centering
    \caption{TabPFN finetuning configuration.}
    \label{tab:tabpfn_config}
    \begin{tabular}{lr}
    \toprule
    \textbf{Hyperparameter} & \textbf{Value} \\
    \midrule
    \texttt{learning\_rate}              & $5 \times 10^{-6}$ \\
    \texttt{epochs}                      & 10 \\
    \texttt{batch\_size}                 & 1024 \\
    \texttt{meta\_batch\_size}           & 1 \\
    \texttt{n\_inference\_context}       & 10\,000 \\
    \texttt{loss}                        & cross-entropy \\
    \bottomrule
    \end{tabular}
\end{table}

In Table~\ref{tab:tabpfn_training_times}, we report the mean finetuning time for TabPFN with prototype undersampling.

\begin{table}[ht]
    \centering
    \caption{Mean training duration (\textit{MM:SS} format) for TabPFN fine-tuning with prototype undersampling across all cross-validation folds.}
    \label{tab:tabpfn_training_times}
    \begin{tabular}{cc}
    \toprule
    \textbf{Horizon} & \textbf{Training time} \\
    \midrule
    0 & 35:43 $\pm$ 00:05 \\
    1 & 31:17 $\pm$ 00:25 \\
    2 & 23:15 $\pm$ 00:09 \\
    3 & 17:40 $\pm$ 00:11 \\
    4 & 13:19 $\pm$ 00:29 \\
    5 & 09:45 $\pm$ 00:04 \\
    \bottomrule
    \end{tabular}
\end{table}

\section{Standard methods hyperparameter configurations}
\label{app:baselines}

Table~\ref{tab:baseline_grids} describes the hyperparameter grids used for standard methods benchmarked against the TabPFN model.

\begin{table}[ht]
    \centering
    \caption{Classical baseline hyperparameter search grids.}
    \label{tab:baseline_grids}
    \begin{tabular}{ll}
    \toprule
    \textbf{Model} & \textbf{Search space} \\
    \midrule
    LR       & $C \in \{10^{-3}, 10^{-2}, 10^{-1}, 1\}$; penalty = L2 \\
    MLP      & hidden sizes $\in \{(64,64),(128,128),(256,256)\}$; \\
             & $\alpha \in \{10^{-4}, 10^{-3}\}$; lr $\in \{10^{-3}, 10^{-2}\}$ \\
    RF       & \texttt{n\_estimators} $\in \{100, 300\}$; \texttt{max\_depth} $\in \{5, 10, \text{None}\}$ \\
    XGBoost  & \texttt{n\_estimators} $\in \{100, 200\}$; \texttt{max\_depth} $\in \{3, 5, 7\}$; \\
             & \texttt{learning\_rate} $\in \{0.05, 0.1, 0.2\}$ \\
    CatBoost & \texttt{iterations} $\in \{100, 200\}$; \texttt{depth} $\in \{4, 6, 8\}$; \\
             & \texttt{learning\_rate} $\in \{0.01, 0.05, 0.1\}$ \\
    LightGBM & \texttt{n\_estimators} $\in \{100, 200\}$; \texttt{max\_depth} $\in \{-1, 5, 10\}$; \\
             & \texttt{learning\_rate} $\in \{0.05, 0.1, 0.2\}$ \\
    \bottomrule
    \end{tabular}
\end{table}

\section{Llama-3-8B finetuning details}
\label{app:llama}

\paragraph{Data formatting.}
Each training instance is formatted as a three-turn conversation with
\texttt{system}, \texttt{user}, and \texttt{assistant} fields.
The system prompt instructs the model to predict bankruptcy for the
specified horizon and respond with a single word: \texttt{YES} or
\texttt{NO}. The user turn contains the company record serialized as
plain key-value pairs (\texttt{column\_name=value}), grouped into named categories (Company Info, Liquidity, Profitability, Leverage, Efficiency,
Growth, Revenue \& Sales, Sector Comparisons, Logarithmic Metrics,
Cash Flow \& Expenses, Equity \& Sales, Risk Flags). Within each group, columns are comma-separated; groups are separated by \texttt{\textbar}. Numeric values are formatted with three decimals, or with thousands separators and two decimals for absolute values $\geq 1000$. Categorical codes for country, region, sector, legal form, employee count, incorporation date, and operational status are rendered as \texttt{code (label)} (e.g., \texttt{country=0 (Poland)}). Missing values are omitted. The assistant turn contains the ground-truth label (\texttt{YES} for distressed, \texttt{NO} for non-distressed).

\paragraph{System prompt.}
\label{app:llama-prompt}

The system prompt was parameterized by the prediction horizon $h$. For each
training and evaluation instance, the placeholder \verb|{horizon_description}| was replaced according to Table~\ref{tab:llama_horizon_prompt_text}.

\begin{quote}
\ttfamily
You are a bankruptcy prediction model.\\
Based on structured company and financial data, predict whether the company
will go bankrupt \{horizon\_description\}.\\
Respond with only one word: YES or NO.
\end{quote}

\begin{table}[ht]
\centering
\caption{Horizon-specific text used in the Llama-3-8B system prompt.}
\label{tab:llama_horizon_prompt_text}
\begin{tabular}{cl}
\toprule
\textbf{Horizon $h$} & \textbf{Replacement for \texttt{\{horizon\_description\}}} \\
\midrule
0 & in the current reporting year \\
1 & one year after the observed reporting year \\
2 & 2 years after the observed reporting year \\
3 & 3 years after the observed reporting year \\
4 & 4 years after the observed reporting year \\
5 & 5 years after the observed reporting year \\
\bottomrule
\end{tabular}
\end{table}

The assistant target was \texttt{YES} for positive examples
(\texttt{Target=1}) and \texttt{NO} for negative examples
(\texttt{Target=0}).

\paragraph{Training configuration.}
Training was conducted on a single NVIDIA A100 GPU using the SLURM workload manager.
The model was loaded in 4-bit precision and only the LoRA adapter matrices were updated.
Key training parameters are summarised in Table~\ref{tab:llama_config}.

\begin{table}[ht]
\centering
\caption{Llama-3-8B QLoRA training configuration.}
\label{tab:llama_config}
\begin{tabular}{lr}
\toprule
\textbf{Hyperparameter} & \textbf{Value} \\
\midrule
\texttt{base\_model}               & \texttt{meta-llama/Meta-Llama-3-8B} \\
\texttt{quantization}              & 4-bit (NF4) \\
\texttt{max\_seq\_length}          & 2048 \\
\texttt{per\_device\_batch\_size}  & 1 \\
\texttt{gradient\_accumulation}    & 8 (effective batch size: 8) \\
\texttt{epochs}                    & 1 \\
\texttt{training\_samples}         & 16{,}000 (80\% of 20{,}000) \\
\texttt{validation\_samples}       & 4{,}000 (20\% of 20{,}000) \\
\bottomrule
\end{tabular}
\end{table}

\paragraph{Evaluation.}
The test set contains all available positive held-out instances for each horizon and a sampled subset of negative held-out instances, up to 100{,}000 total observations. This preserves all minority-class test examples available under the split while keeping evaluation computationally feasible. The resulting class distribution is therefore not the natural V4FinBench class distribution and should be interpreted as a controlled, positive-enriched diagnostic evaluation. For hard-label evaluation, metrics are computed only on instances where the model produced a valid \texttt{YES} or \texttt{NO} response; cases where the model generated other output are counted as unknown predictions and excluded. For probability-based evaluation, we extract token-level log-probabilities for the \texttt{YES} and \texttt{NO} tokens to obtain soft probability estimates, allowing threshold calibration and ROC-AUC computation on all instances.

\section{Full results}
\label{app:full-results}

Table~\ref{tab:complete_all_results} gathers results of standard methods and three variants of finetuned TabPFN across all prediction horizons and metrics.

\begin{tiny}
\begin{longtable}{llccccc}
\caption{Performance results across all prediction horizons and metrics.}
\label{tab:complete_all_results}\\
\toprule
H & Model & Accuracy & Precision & Recall & $F_1$ & ROC-AUC \\
\midrule
\endfirsthead

\multicolumn{7}{c}{{\bfseries Table \thetable\ continued from previous page}} \\
\toprule
H & Model & Accuracy & Precision & Recall & $F_1$ & ROC-AUC \\
\midrule
\endhead

\midrule
\multicolumn{7}{r}{{Continued on next page}} \\
\endfoot

\bottomrule
\endlastfoot

\multirow{9}{*}{0}
& XGBoost & 0.997 $\pm$ 0.000 & 0.387 $\pm$ 0.021 & 0.587 $\pm$ 0.053 & 0.465 $\pm$ 0.004 & 0.996 $\pm$ 0.000 \\
& CatBoost & 0.997 $\pm$ 0.000 & 0.368 $\pm$ 0.030 & 0.577 $\pm$ 0.060 & 0.446 $\pm$ 0.006 & 0.996 $\pm$ 0.000 \\
& LightGBM & 0.997 $\pm$ 0.000 & 0.381 $\pm$ 0.024 & 0.588 $\pm$ 0.031 & 0.461 $\pm$ 0.011 & 0.996 $\pm$ 0.000 \\
& MLP & 0.996 $\pm$ 0.000 & 0.200 $\pm$ 0.022 & 0.367 $\pm$ 0.041 & 0.258 $\pm$ 0.018 & 0.975 $\pm$ 0.005 \\
& RandomForest & 0.986 $\pm$ 0.001 & 0.241 $\pm$ 0.005 & 0.675 $\pm$ 0.067 & 0.355 $\pm$ 0.011 & 0.994 $\pm$ 0.000 \\
& LR & 0.996 $\pm$ 0.000 & 0.193 $\pm$ 0.019 & 0.378 $\pm$ 0.066 & 0.252 $\pm$ 0.011 & 0.983 $\pm$ 0.001 \\
& TabPFN (none) & 0.997 $\pm$ 0.000 & 0.204 $\pm$ 0.007 & 0.713 $\pm$ 0.069 & 0.317 $\pm$ 0.010 & 0.993 $\pm$ 0.001 \\
& TabPFN (prototype) & 0.982 $\pm$ 0.003 & 0.290 $\pm$ 0.031 & 0.685 $\pm$ 0.073 & 0.405 $\pm$ 0.024 & 0.995 $\pm$ 0.000 \\
& TabPFN (undersample) & 0.983 $\pm$ 0.002 & 0.273 $\pm$ 0.030 & 0.667 $\pm$ 0.061 & 0.386 $\pm$ 0.028 & 0.995 $\pm$ 0.001 \\
\midrule

\multirow{9}{*}{1}
& XGBoost & 0.997 $\pm$ 0.000 & 0.139 $\pm$ 0.005 & 0.254 $\pm$ 0.044 & 0.179 $\pm$ 0.011 & 0.941 $\pm$ 0.005 \\
& CatBoost & 0.997 $\pm$ 0.000 & 0.137 $\pm$ 0.011 & 0.250 $\pm$ 0.043 & 0.175 $\pm$ 0.009 & 0.937 $\pm$ 0.002 \\
& LightGBM & 0.997 $\pm$ 0.000 & 0.139 $\pm$ 0.013 & 0.262 $\pm$ 0.041 & 0.181 $\pm$ 0.013 & 0.931 $\pm$ 0.009 \\
& MLP & 0.997 $\pm$ 0.000 & 0.071 $\pm$ 0.012 & 0.213 $\pm$ 0.052 & 0.105 $\pm$ 0.017 & 0.901 $\pm$ 0.010 \\
& RandomForest & 0.973 $\pm$ 0.001 & 0.095 $\pm$ 0.016 & 0.231 $\pm$ 0.064 & 0.130 $\pm$ 0.014 & 0.918 $\pm$ 0.010 \\
& LR & 0.997 $\pm$ 0.000 & 0.067 $\pm$ 0.007 & 0.161 $\pm$ 0.027 & 0.093 $\pm$ 0.007 & 0.903 $\pm$ 0.006 \\
& TabPFN (none) & 0.997 $\pm$ 0.000 & 0.080 $\pm$ 0.012 & 0.224 $\pm$ 0.038 & 0.116 $\pm$ 0.008 & 0.911 $\pm$ 0.005 \\
& TabPFN (prototype) & 0.947 $\pm$ 0.007 & 0.130 $\pm$ 0.015 & 0.246 $\pm$ 0.043 & 0.169 $\pm$ 0.012 & 0.944 $\pm$ 0.005 \\
& TabPFN (undersample) & 0.945 $\pm$ 0.010 & 0.126 $\pm$ 0.012 & 0.223 $\pm$ 0.030 & 0.160 $\pm$ 0.011 & 0.938 $\pm$ 0.005 \\
\midrule

\multirow{9}{*}{2}
& XGBoost & 0.997 $\pm$ 0.000 & 0.078 $\pm$ 0.006 & 0.146 $\pm$ 0.057 & 0.098 $\pm$ 0.018 & 0.902 $\pm$ 0.007 \\
& CatBoost & 0.997 $\pm$ 0.000 & 0.081 $\pm$ 0.007 & 0.126 $\pm$ 0.048 & 0.096 $\pm$ 0.017 & 0.897 $\pm$ 0.006 \\
& LightGBM & 0.997 $\pm$ 0.000 & 0.076 $\pm$ 0.005 & 0.129 $\pm$ 0.038 & 0.094 $\pm$ 0.013 & 0.895 $\pm$ 0.006 \\
& MLP & 0.997 $\pm$ 0.000 & 0.035 $\pm$ 0.005 & 0.115 $\pm$ 0.025 & 0.053 $\pm$ 0.006 & 0.834 $\pm$ 0.018 \\
& RandomForest & 0.975 $\pm$ 0.001 & 0.055 $\pm$ 0.013 & 0.120 $\pm$ 0.049 & 0.069 $\pm$ 0.008 & 0.874 $\pm$ 0.008 \\
& LR & 0.997 $\pm$ 0.000 & 0.040 $\pm$ 0.004 & 0.108 $\pm$ 0.035 & 0.058 $\pm$ 0.006 & 0.854 $\pm$ 0.011 \\
& TabPFN (none) & 0.997 $\pm$ 0.000 & 0.043 $\pm$ 0.012 & 0.153 $\pm$ 0.058 & 0.064 $\pm$ 0.014 & 0.861 $\pm$ 0.013 \\
& TabPFN (prototype) & 0.926 $\pm$ 0.010 & 0.083 $\pm$ 0.005 & 0.160 $\pm$ 0.022 & 0.108 $\pm$ 0.007 & 0.908 $\pm$ 0.005 \\
& TabPFN (undersample) & 0.936 $\pm$ 0.010 & 0.069 $\pm$ 0.008 & 0.161 $\pm$ 0.039 & 0.095 $\pm$ 0.011 & 0.900 $\pm$ 0.006 \\
\midrule

\multirow{9}{*}{3}
& XGBoost & 0.998 $\pm$ 0.000 & 0.047 $\pm$ 0.005 & 0.126 $\pm$ 0.029 & 0.067 $\pm$ 0.001 & 0.878 $\pm$ 0.009 \\
& CatBoost & 0.998 $\pm$ 0.000 & 0.053 $\pm$ 0.011 & 0.113 $\pm$ 0.077 & 0.064 $\pm$ 0.011 & 0.876 $\pm$ 0.011 \\
& LightGBM & 0.998 $\pm$ 0.000 & 0.043 $\pm$ 0.003 & 0.120 $\pm$ 0.033 & 0.063 $\pm$ 0.007 & 0.874 $\pm$ 0.010 \\
& MLP & 0.998 $\pm$ 0.000 & 0.032 $\pm$ 0.006 & 0.062 $\pm$ 0.026 & 0.041 $\pm$ 0.007 & 0.811 $\pm$ 0.007 \\
& RandomForest & 0.979 $\pm$ 0.002 & 0.035 $\pm$ 0.006 & 0.122 $\pm$ 0.033 & 0.053 $\pm$ 0.008 & 0.848 $\pm$ 0.011 \\
& LR & 0.998 $\pm$ 0.000 & 0.028 $\pm$ 0.003 & 0.086 $\pm$ 0.035 & 0.041 $\pm$ 0.007 & 0.826 $\pm$ 0.011 \\
& TabPFN (none) & 0.998 $\pm$ 0.000 & 0.029 $\pm$ 0.020 & 0.095 $\pm$ 0.068 & 0.038 $\pm$ 0.013 & 0.834 $\pm$ 0.015 \\
& TabPFN (prototype) & 0.923 $\pm$ 0.010 & 0.054 $\pm$ 0.013 & 0.140 $\pm$ 0.066 & 0.073 $\pm$ 0.010 & 0.888 $\pm$ 0.009 \\
& TabPFN (undersample) & 0.922 $\pm$ 0.006 & 0.047 $\pm$ 0.011 & 0.120 $\pm$ 0.075 & 0.060 $\pm$ 0.011 & 0.877 $\pm$ 0.007 \\
\midrule

\multirow{9}{*}{4}
& XGBoost & 0.998 $\pm$ 0.000 & 0.037 $\pm$ 0.004 & 0.106 $\pm$ 0.039 & 0.053 $\pm$ 0.006 & 0.852 $\pm$ 0.011 \\
& CatBoost & 0.998 $\pm$ 0.000 & 0.042 $\pm$ 0.007 & 0.094 $\pm$ 0.037 & 0.055 $\pm$ 0.008 & 0.859 $\pm$ 0.016 \\
& LightGBM & 0.998 $\pm$ 0.000 & 0.038 $\pm$ 0.008 & 0.094 $\pm$ 0.014 & 0.053 $\pm$ 0.009 & 0.854 $\pm$ 0.013 \\
& MLP & 0.998 $\pm$ 0.000 & 0.019 $\pm$ 0.004 & 0.049 $\pm$ 0.025 & 0.026 $\pm$ 0.007 & 0.754 $\pm$ 0.027 \\
& RandomForest & 0.986 $\pm$ 0.002 & 0.028 $\pm$ 0.010 & 0.079 $\pm$ 0.041 & 0.037 $\pm$ 0.008 & 0.825 $\pm$ 0.022 \\
& LR & 0.998 $\pm$ 0.000 & 0.019 $\pm$ 0.004 & 0.058 $\pm$ 0.046 & 0.026 $\pm$ 0.008 & 0.805 $\pm$ 0.021 \\
& TabPFN (none) & 0.998 $\pm$ 0.000 & 0.016 $\pm$ 0.005 & 0.107 $\pm$ 0.108 & 0.022 $\pm$ 0.007 & 0.786 $\pm$ 0.022 \\
& TabPFN (prototype) & 0.918 $\pm$ 0.009 & 0.046 $\pm$ 0.010 & 0.086 $\pm$ 0.039 & 0.055 $\pm$ 0.008 & 0.870 $\pm$ 0.006 \\
& TabPFN (undersample) & 0.917 $\pm$ 0.014 & 0.029 $\pm$ 0.009 & 0.111 $\pm$ 0.033 & 0.045 $\pm$ 0.012 & 0.863 $\pm$ 0.007 \\
\midrule

\multirow{9}{*}{5}
& XGBoost & 0.998 $\pm$ 0.000 & 0.037 $\pm$ 0.008 & 0.071 $\pm$ 0.042 & 0.045 $\pm$ 0.013 & 0.837 $\pm$ 0.007 \\
& CatBoost & 0.998 $\pm$ 0.000 & 0.034 $\pm$ 0.008 & 0.071 $\pm$ 0.032 & 0.044 $\pm$ 0.010 & 0.841 $\pm$ 0.008 \\
& LightGBM & 0.998 $\pm$ 0.000 & 0.037 $\pm$ 0.008 & 0.079 $\pm$ 0.007 & 0.050 $\pm$ 0.008 & 0.826 $\pm$ 0.009 \\
& MLP & 0.998 $\pm$ 0.000 & 0.023 $\pm$ 0.011 & 0.042 $\pm$ 0.020 & 0.027 $\pm$ 0.010 & 0.745 $\pm$ 0.026 \\
& RandomForest & 0.991 $\pm$ 0.001 & 0.021 $\pm$ 0.006 & 0.065 $\pm$ 0.062 & 0.026 $\pm$ 0.005 & 0.796 $\pm$ 0.008 \\
& LR & 0.998 $\pm$ 0.000 & 0.021 $\pm$ 0.005 & 0.063 $\pm$ 0.060 & 0.028 $\pm$ 0.008 & 0.793 $\pm$ 0.007 \\
& TabPFN (none) & 0.998 $\pm$ 0.000 & 0.015 $\pm$ 0.004 & 0.081 $\pm$ 0.061 & 0.024 $\pm$ 0.005 & 0.780 $\pm$ 0.020 \\
& TabPFN (prototype) & 0.922 $\pm$ 0.008 & 0.045 $\pm$ 0.019 & 0.087 $\pm$ 0.044 & 0.052 $\pm$ 0.015 & 0.860 $\pm$ 0.010 \\
& TabPFN (undersample) & 0.931 $\pm$ 0.007 & 0.036 $\pm$ 0.011 & 0.082 $\pm$ 0.028 & 0.048 $\pm$ 0.010 & 0.846 $\pm$ 0.013 \\
\end{longtable}
\end{tiny}

\section{Data}

\subsection{Data source and preparation}
\label{app:data-source}

V4FinBench is constructed from company-level financial statement data sourced from the Emerging Markets Information Service (EMIS) database. We extracted raw financial-statement items and company metadata, including country, legal form, industry classification, operational status, incorporation information, and number of employees.

The financial indicators used in V4FinBench were computed from raw accounting items rather than taken as precomputed ratios. The raw statement fields include, among others, net profit, sales revenue, total operating revenue, total assets, fixed assets, equity, current assets, current liabilities, inventories, receivables, EBITDA, EBIT, interest expense, cash and cash equivalents, operating cash flow, total liabilities, and non-current liabilities. During preprocessing, numeric statement fields were converted to numeric values, records with insufficient information in essential accounting fields were removed, and infinite values produced by ratio calculations were treated as missing.

From these raw items, we derived financial ratios and indicators covering liquidity, profitability, solvency, turnover and operating-cycle measures, growth measures, firm-size transformations, and binary risk flags. We also constructed sector-relative indicators by first computing company-level ratios and corresponding sector-level aggregates, and then comparing each company to its sector benchmark. Industry information was additionally encoded using NAICS-derived features, including primary and secondary NAICS codes, 2-digit and 3-digit NAICS groupings, and an indicator for firms with multiple industries.

V4FinBench covers companies registered in the four Visegrád Group economies: Poland, Hungary, Czech Republic, and Slovakia. The dataset spans 2006--2021, a period encompassing the 2008--2009 global financial crisis, the European sovereign debt crisis, and the COVID-19 pandemic. In total, the benchmark contains 203{,}900 unique companies and 1{,}106{,}879 company-year observations, distributed as Poland (628{,}499), Hungary (358{,}486), Slovakia (62{,}141), and Czech Republic (57{,}753). The complete list of released features is provided in Appendix~\ref{app:features}.

\subsection{Composite financial distress definition}
\label{app:distress-definition}

\paragraph{Comparison with prior distress definitions in the V4 literature.}
Most prior bankruptcy-prediction studies in the Visegr\'{a}d region rely on
formal bankruptcy decisions (declarations, filings, or ongoing proceedings) \cite{tomczak2025identification},
or on partial single-indicator criteria. Slovak studies frequently follow
the ``enterprise in crisis'' definition from \S3 of Act No.~7/2005~Coll.\ on
Bankruptcy and Restructuring, often supplemented by an equity-to-liability
ratio below 0.08~\cite{Durica_Frnda_Svabova_2023,Gavurova_Jencova_Bacik_Miskufova_Letkovsky_2022,Valaskova_Gajdosikova_Belas_2023}.
Other works define distress as simultaneously negative EBITDA, EBIT, and
net profit~\cite{geise2021corporate,platt2006comparing}, while
\cite{tomczak2025} relies solely on negative equity. Our composite
definition is more restrictive: by jointly requiring deterioration in 
solvency, profitability, and liquidity, it isolates companies in
genuine financial distress rather than those exhibiting weakness in a
single dimension.

\subsection{Feature set}
\label{app:features}

Each company-year observation is described by 131 features encompassing
seven categories: company metadata (e.g., country, legal form, industry
classification), liquidity and profitability ratios, turnover and cycle
ratios, solvency ratios, year-over-year growth rates, company size measures
(logarithmic transformations), and sector-relative indicators that capture
a company's deviation from its industry-year average.
Table~\ref{tab:financial_indicators} provides the complete list of all
131 variables with descriptions.

{\footnotesize
\setlength{\tabcolsep}{4pt}
\begin{longtable}{@{} l p{5.4cm} @{\hspace{0.6cm}} l p{5.4cm} @{}}
\caption{Complete list of 131 financial and non-financial indicators used as input features.}
\label{tab:financial_indicators} \\
\toprule
\textbf{Var.} & \textbf{Description} & \textbf{Var.} & \textbf{Description} \\
\midrule
\endfirsthead

\multicolumn{4}{c}%
{{\bfseries \tablename\ \thetable{} -- continued from previous page}} \\
\toprule
\textbf{Var.} & \textbf{Description} & \textbf{Var.} & \textbf{Description} \\
\midrule
\endhead

\midrule
\multicolumn{4}{r}{{Continued on next page}} \\
\endfoot
\endlastfoot

\multicolumn{4}{l}{\textbf{Company Identifiers \& Metadata}} \\
\midrule
X1 & country & X2 & has\_multiple\_industries \\
X3 & incorporation\_date\_1 & X4 & incorporation\_date\_2 \\
X5 & legal\_form & X6 & naics\_2digit \\
X7 & naics\_3digit & X8 & number\_of\_employees \\
X9 & operational\_status & X10 & primary\_naics\_encoded \\
X11 & secondary\_naics\_encoded & X12 & sector\_1 \\
X13 & state & X14 & year \\
\midrule

\multicolumn{4}{l}{\textbf{Liquidity and Profitability Ratios}} \\
\midrule
X15 & Cash/sales & X16 & Cash/total\_assets \\
X17 & Cash/total\_operating\_revenue & X18 & Current\_assets-inventories-receivables/short\_term\_liabilities \\
X19 & Current\_assets-inventories/short\_term\_liabilities & X20 & Current\_assets/sales \\
X21 & Current\_assets/short\_term\_liabilities & X22 & EBIT/equity \\
X23 & EBIT/financial\_costs & X24 & EBIT/total\_assets \\
X25 & EBIT/total\_costs & X26 & EBIT/total\_liabilities \\
X27 & EBIT/total\_operating\_revenue & X28 & EBITDA/fixed\_assets \\
X29 & EBITDA/total\_assets & X30 & EBITDA/total\_operating\_revenue \\
X31 & Gross\_profit+depreciation/total\_liabilities & X32 & Gross\_profit/short\_term\_liabilities \\
X33 & Gross\_profit/total\_assets & X34 & Gross\_profit/total\_operating\_revenue \\
X35 & Interest\_expense/revenue & X36 & Inventories/working\_capital \\
X37 & Net\_profit+depreciation/current\_liabilities & X38 & Net\_profit+depreciation/total\_liabilities \\
X39 & Net\_profit/equity & X40 & Net\_profit/fixed\_assets \\
X41 & Net\_profit/inventories & X42 & Net\_profit/total\_assets \\
X43 & Net\_profit/total\_operating\_revenue & X44 & Net\_profit/current\_assets \\
X45 & Operational\_expenses/short\_term\_liabilities & X46 & Operational\_expenses/total\_liabilities \\
X47 & Quick\_assets/total\_operating\_revenue & X48 & Retained\_profit/short\_term\_liabilities \\
X49 & Retained\_profit/total\_assets & X50 & Working\_capital \\
X51 & Working\_capital/equity & X52 & Working\_capital/fixed\_assets \\
X53 & Working\_capital/sales & X54 & Working\_capital/total\_assets \\
X55 & Working\_capital/total\_liabilities & X56 & Working\_capital/total\_operating\_revenue \\
X57 & Cash\_flow/total\_operating\_revenue & X58 & Cash\_flow/total\_debt \\
X59 & Loss\_flag (Net Profit (Loss) < 0) & & \\
\midrule

\multicolumn{4}{l}{\textbf{Turnover and Cycle Ratios}} \\
\midrule
X60 & Cash\_conversion\_cycle & X61 & Inventories/total\_operating\_revenue \\
X62 & Operating\_cycle & X63 & Operating\_expenses/total\_operating\_revenue \\
X64 & Receivables\_turnover\_days & X65 & Revenue/current\_assets \\
X66 & Revenue/long\_term\_liabilities & X67 & Revenue/total\_liabilities \\
X68 & Short\_term\_liabilities\_turnover\_days & X69 & Total\_operating\_revenue/fixed\_assets \\
X70 & Total\_operating\_revenue/inventories & X71 & Total\_operating\_revenue/receivables \\
X72 & Total\_operating\_revenue/short\_term\_liab. & X73 & Total\_operating\_revenue/total\_assets \\
\midrule

\multicolumn{4}{l}{\textbf{Solvency Ratios}} \\
\midrule
X74 & Constant\_capital/fixed\_assets & X75 & Constant\_capital/total\_assets \\
X76 & Current\_assets/total\_liabilities & X77 & Current\_assets/total\_operating\_revenue \\
X78 & Current\_liabilities/total\_liabilities & X79 & Current\_liabilities/current\_assets \\
X80 & Current\_liabilities/equity & X81 & Short\_term\_liabilities/total\_assets \\
X82 & Equity-share\_capital/fixed\_assets & X83 & Equity/fixed\_assets \\
X84 & Equity/long\_term\_liabilities & X85 & Equity/total\_operating\_revenue \\
X86 & Equity/total\_assets & X87 & Equity/total\_liabilities \\
X88 & Equity\_ratio\_classification & X89 & Fixed\_assets/long\_term\_liabilities \\
X90 & Fixed\_assets/total\_assets & X91 & Inventories+receivables/equity \\
X92 & Inventory/current\_liabilities & X93 & Long\_term\_liabilities/current\_assets \\
X94 & Long\_term\_liabilities/equity & X95 & Total\_liabilities-cash/EBITDA \\
X96 & Total\_liabilities-cash/total\_operating\_rev & X97 & Total\_liabilities/total\_assets \\
X98 & Insolvency\_flag (Total\_liab > Total\_assets) & & \\
\midrule

\multicolumn{4}{l}{\textbf{Growth Ratios (YoY)}} \\
\midrule
X99 & Current\_assets\_growth & X100 & Inventories\_growth \\
X101 & Net\_profit\_growth & X102 & Operating\_profit\_growth \\
X103 & Operating\_revenue\_growth & X104 & Receivables\_growth \\
X105 & Short\_term\_liabilities\_growth & X106 & Total\_assets\_growth \\
\midrule

\multicolumn{4}{l}{\textbf{Company Size Ratios}} \\
\midrule
X107 & Log\_current\_assets & X108 & Log\_net\_profit/gdp \\
X109 & Log\_operating\_profit/gdp & X110 & Log\_revenue/gdp \\
X111 & Log\_total\_liabilities & X112 & Logarithm\_of\_total\_assets \\
X113 & Logarithm\_of\_total\_assets/GDP & X114 & Logarithm\_of\_total\_operating\_revenue \\
\midrule

\multicolumn{4}{l}{\textbf{Sector-Relative Indicators}} \\
\multicolumn{4}{p{13cm}}{\textit{Features represent the difference between the company's ratio and the industry sector average for that year.}} \\
\midrule
X115 & Cash\_conversion\_cycle\_Sector & X116 & Current\_liabilities*365/revenue\_Sector \\
X117 & Current\_assets/current\_liabilities\_Sector & X118 & EBITDA\_margin\_Sector \\
X119 & Inventories*365/revenue\_Sector & X120 & Net\_profit/abs\_equity\_Sector \\
X121 & Net\_profit/assets\_Sector & X122 & Net\_profit/current\_assets\_Sector \\
X123 & Net\_profit/fixed\_assets\_Sector & X124 & Net\_profit/total\_operating\_revenue\_Sector \\
X125 & Operating\_cycle\_Sector & X126 & Receivables*365/revenue\_Sector \\
X127 & Revenue/assets\_Sector & X128 & Revenue/fixed\_assets\_Sector \\
X129 & ST\_financial\_assets/current\_liabilities\_Sec & X130 & ST\_receivables\_investments/current\_liab\_Sec \\
X131 & Working\_capital/assets\_Sector & & \\
\bottomrule

\end{longtable}}

\section{Licenses for existing assets}
\label{app:licenses}

\paragraph{Data sources.}
V4FinBench is built from data acquired from the Emerging Markets Information Service (EMIS, \url{https://www.emis.com/}) under an individual data-purchase agreement. EMIS was informed that the data was collected as part of a research grant and would be publicly released as part of the resulting benchmark, and granted explicit permission for
this release. The released artifact consists of derived financial indicators and aggregated company-year records computed from raw statement items; proprietary EMIS identifiers are not redistributed.
The American Bankruptcy Dataset~\cite{lombardo2022machine}, used in
the transfer experiment (Section~\ref{sec:exp-transfer}), is released under \href{https://creativecommons.org/publicdomain/zero/1.0/}{CC0~1.0 (Public Domain Dedication)}.

\paragraph{Models and libraries.}
We use \texttt{meta-llama/Meta-Llama-3-8B}~\cite{llama3_modelcard} under the Meta Llama~3 Community License Agreement, and TabPFN v2~\cite{hollmann2025tabpfn} under the license released by its authors. Classical baselines are implemented with open-source libraries: XGBoost~\cite{chen2016xgboost} (Apache~2.0),
CatBoost~\cite{prokhorenkova2018catboost} (Apache~2.0), LightGBM~\cite{ke2017lightgbm} (MIT), and scikit-learn (BSD-3-Clause) for logistic regression, multilayer perceptron, and random forest. All assets are used in accordance with their respective licenses, and attribution is provided through the bibliography.

\paragraph{Released artifacts.}
The V4FinBench dataset is published on Kaggle
(\url{https://www.kaggle.com/datasets/sebastiantomczak10/v4-group-corporate-bankruptcy/data}),
the label-construction code on GitHub
(\url{https://github.com/genwro-ai/V4FinBench}), and the finetuned
TabPFN checkpoints on Hugging Face
(\url{https://huggingface.co/Manik2000/v4finbench-tabpfn}).

\newpage
\section*{NeurIPS Paper Checklist}

\begin{enumerate}

\item {\bf Claims}
    \item[] Question: Do the main claims made in the abstract and introduction accurately reflect the paper's contributions and scope?
    \item[] Answer: \answerYes{} 
    \item[] Justification: The abstract and introduction clearly state the paper's main contributions: the release of V4FinBench, reference evaluations of standard baselines, TabPFN, and Llama-3-8B, and a cross-dataset transfer analysis. These claims are supported by the dataset description in Section~\ref{sec:dataset}, the evaluation protocol in Section~\ref{sec:eval-protocol}, the experiments in Section~\ref{sec:experiments}.
    \item[] Guidelines:
    \begin{itemize}
        \item The answer \answerNA{} means that the abstract and introduction do not include the claims made in the paper.
        \item The abstract and/or introduction should clearly state the claims made, including the contributions made in the paper and important assumptions and limitations. A \answerNo{} or \answerNA{} answer to this question will not be perceived well by the reviewers. 
        \item The claims made should match theoretical and experimental results, and reflect how much the results can be expected to generalize to other settings. 
        \item It is fine to include aspirational goals as motivation as long as it is clear that these goals are not attained by the paper. 
    \end{itemize}

\item {\bf Limitations}
    \item[] Question: Does the paper discuss the limitations of the work performed by the authors?
    \item[] Answer: \answerYes{} 
    \item[] Justification: The paper includes a dedicated Limitations section, Section~\ref{sec:limitations}, discussing geographic scope, the use of composite financial-distress labels rather than formal legal bankruptcy filings, and responsible-use concerns for downstream deployment. It also clarifies that V4FinBench is intended for benchmarking and research rather than automated decision-making.
    \item[] Guidelines:
    \begin{itemize}
        \item The answer \answerNA{} means that the paper has no limitation while the answer \answerNo{} means that the paper has limitations, but those are not discussed in the paper. 
        \item The authors are encouraged to create a separate ``Limitations'' section in their paper.
        \item The paper should point out any strong assumptions and how robust the results are to violations of these assumptions (e.g., independence assumptions, noiseless settings, model well-specification, asymptotic approximations only holding locally). The authors should reflect on how these assumptions might be violated in practice and what the implications would be.
        \item The authors should reflect on the scope of the claims made, e.g., if the approach was only tested on a few datasets or with a few runs. In general, empirical results often depend on implicit assumptions, which should be articulated.
        \item The authors should reflect on the factors that influence the performance of the approach. For example, a facial recognition algorithm may perform poorly when image resolution is low or images are taken in low lighting. Or a speech-to-text system might not be used reliably to provide closed captions for online lectures because it fails to handle technical jargon.
        \item The authors should discuss the computational efficiency of the proposed algorithms and how they scale with dataset size.
        \item If applicable, the authors should discuss possible limitations of their approach to address problems of privacy and fairness.
        \item While the authors might fear that complete honesty about limitations might be used by reviewers as grounds for rejection, a worse outcome might be that reviewers discover limitations that aren't acknowledged in the paper. The authors should use their best judgment and recognize that individual actions in favor of transparency play an important role in developing norms that preserve the integrity of the community. Reviewers will be specifically instructed to not penalize honesty concerning limitations.
    \end{itemize}

\item {\bf Theory assumptions and proofs}
    \item[] Question: For each theoretical result, does the paper provide the full set of assumptions and a complete (and correct) proof?
    \item[] Answer: \answerNA{} 
    \item[] Justification: The paper does not present theoretical results, theorems, or proofs. Its contributions are empirical and dataset-oriented, focusing on benchmark construction, evaluation protocols, and experimental comparisons.
    \item[] Guidelines:
    \begin{itemize}
        \item The answer \answerNA{} means that the paper does not include theoretical results. 
        \item All the theorems, formulas, and proofs in the paper should be numbered and cross-referenced.
        \item All assumptions should be clearly stated or referenced in the statement of any theorems.
        \item The proofs can either appear in the main paper or the supplemental material, but if they appear in the supplemental material, the authors are encouraged to provide a short proof sketch to provide intuition. 
        \item Inversely, any informal proof provided in the core of the paper should be complemented by formal proofs provided in appendix or supplemental material.
        \item Theorems and Lemmas that the proof relies upon should be properly referenced. 
    \end{itemize}

    \item {\bf Experimental result reproducibility}
    \item[] Question: Does the paper fully disclose all the information needed to reproduce the main experimental results of the paper to the extent that it affects the main claims and/or conclusions of the paper (regardless of whether the code and data are provided or not)?
    \item[] Answer: \answerYes{} 
    \item[] Justification: The paper specifies the dataset construction, released fold-based evaluation protocol, preprocessing, metrics, threshold calibration, model families, hyperparameter grids, and finetuning details in Sections~\ref{sec:dataset}--\ref{sec:methods}, Section~\ref{sec:experiments}, and Appendices~\ref{app:tabpfn}--\ref{app:llama}. The dataset, label-derivation code, and TabPFN checkpoints are also linked in the paper.
    \item[] Guidelines:
    \begin{itemize}
        \item The answer \answerNA{} means that the paper does not include experiments.
        \item If the paper includes experiments, a \answerNo{} answer to this question will not be perceived well by the reviewers: Making the paper reproducible is important, regardless of whether the code and data are provided or not.
        \item If the contribution is a dataset and\slash or model, the authors should describe the steps taken to make their results reproducible or verifiable. 
        \item Depending on the contribution, reproducibility can be accomplished in various ways. For example, if the contribution is a novel architecture, describing the architecture fully might suffice, or if the contribution is a specific model and empirical evaluation, it may be necessary to either make it possible for others to replicate the model with the same dataset, or provide access to the model. In general. releasing code and data is often one good way to accomplish this, but reproducibility can also be provided via detailed instructions for how to replicate the results, access to a hosted model (e.g., in the case of a large language model), releasing of a model checkpoint, or other means that are appropriate to the research performed.
        \item While NeurIPS does not require releasing code, the conference does require all submissions to provide some reasonable avenue for reproducibility, which may depend on the nature of the contribution. For example
        \begin{enumerate}
            \item If the contribution is primarily a new algorithm, the paper should make it clear how to reproduce that algorithm.
            \item If the contribution is primarily a new model architecture, the paper should describe the architecture clearly and fully.
            \item If the contribution is a new model (e.g., a large language model), then there should either be a way to access this model for reproducing the results or a way to reproduce the model (e.g., with an open-source dataset or instructions for how to construct the dataset).
            \item We recognize that reproducibility may be tricky in some cases, in which case authors are welcome to describe the particular way they provide for reproducibility. In the case of closed-source models, it may be that access to the model is limited in some way (e.g., to registered users), but it should be possible for other researchers to have some path to reproducing or verifying the results.
        \end{enumerate}
    \end{itemize}

\item {\bf Open access to data and code}
    \item[] Question: Does the paper provide open access to the data and code, with sufficient instructions to faithfully reproduce the main experimental results, as described in supplemental material?
    \item[] Answer: \answerYes{} 
    \item[] Justification: The paper provides public links to the V4FinBench dataset on Kaggle, the label-derivation code on GitHub, and the best TabPFN checkpoints on Hugging Face. These resources, together with the evaluation protocol, preprocessing details, hyperparameter grids, and finetuning configurations described in the paper and appendix, allow for reproducing the main results.
    \item[] Guidelines:
    \begin{itemize}
        \item The answer \answerNA{} means that paper does not include experiments requiring code.
        \item Please see the NeurIPS code and data submission guidelines (\url{https://neurips.cc/public/guides/CodeSubmissionPolicy}) for more details.
        \item While we encourage the release of code and data, we understand that this might not be possible, so \answerNo{} is an acceptable answer. Papers cannot be rejected simply for not including code, unless this is central to the contribution (e.g., for a new open-source benchmark).
        \item The instructions should contain the exact command and environment needed to run to reproduce the results. See the NeurIPS code and data submission guidelines (\url{https://neurips.cc/public/guides/CodeSubmissionPolicy}) for more details.
        \item The authors should provide instructions on data access and preparation, including how to access the raw data, preprocessed data, intermediate data, and generated data, etc.
        \item The authors should provide scripts to reproduce all experimental results for the new proposed method and baselines. If only a subset of experiments are reproducible, they should state which ones are omitted from the script and why.
        \item At submission time, to preserve anonymity, the authors should release anonymized versions (if applicable).
        \item Providing as much information as possible in supplemental material (appended to the paper) is recommended, but including URLs to data and code is permitted.
    \end{itemize}

\item {\bf Experimental setting/details}
    \item[] Question: Does the paper specify all the training and test details (e.g., data splits, hyperparameters, how they were chosen, type of optimizer) necessary to understand the results?
    \item[] Answer: \answerYes{} 
    \item[] Justification: The paper describes the 5-fold grouped stratified evaluation protocol, train/validation/test split logic, preprocessing, metrics, threshold calibration, and hyperparameter selection in Sections~\ref{sec:eval-protocol} and~\ref{sec:experiments}. Additional method-specific details, including the TabPFN optimizer and finetuning configuration, classical baseline grids, and Llama-3-8B QLoRA setup, are provided in Appendices~\ref{app:tabpfn},~\ref{app:baselines}, and~\ref{app:llama}.
    \item[] Guidelines:
    \begin{itemize}
        \item The answer \answerNA{} means that the paper does not include experiments.
        \item The experimental setting should be presented in the core of the paper to a level of detail that is necessary to appreciate the results and make sense of them.
        \item The full details can be provided either with the code, in appendix, or as supplemental material.
    \end{itemize}

\item {\bf Experiment statistical significance}
    \item[] Question: Does the paper report error bars suitably and correctly defined or other appropriate information about the statistical significance of the experiments?
    \item[] Answer: \answerYes{} 
    \item[] Justification: The main TabPFN and standard-baseline experiments report mean performance with standard deviations across the five cross-validation folds, as shown in Figure~\ref{fig:tabpfn-vs-baselines} and Table~\ref{tab:complete_all_results}. The figure caption explicitly states that shaded bands correspond to $\pm 1$ standard deviation across folds.
    \item[] Guidelines:
    \begin{itemize}
        \item The answer \answerNA{} means that the paper does not include experiments.
        \item The authors should answer \answerYes{} if the results are accompanied by error bars, confidence intervals, or statistical significance tests, at least for the experiments that support the main claims of the paper.
        \item The factors of variability that the error bars are capturing should be clearly stated (for example, train/test split, initialization, random drawing of some parameter, or overall run with given experimental conditions).
        \item The method for calculating the error bars should be explained (closed form formula, call to a library function, bootstrap, etc.)
        \item The assumptions made should be given (e.g., Normally distributed errors).
        \item It should be clear whether the error bar is the standard deviation or the standard error of the mean.
        \item It is OK to report 1-sigma error bars, but one should state it. The authors should preferably report a 2-sigma error bar than state that they have a 96\% CI, if the hypothesis of Normality of errors is not verified.
        \item For asymmetric distributions, the authors should be careful not to show in tables or figures symmetric error bars that would yield results that are out of range (e.g., negative error rates).
        \item If error bars are reported in tables or plots, the authors should explain in the text how they were calculated and reference the corresponding figures or tables in the text.
    \end{itemize}

\item {\bf Experiments compute resources}
    \item[] Question: For each experiment, does the paper provide sufficient information on the computer resources (type of compute workers, memory, time of execution) needed to reproduce the experiments?
    \item[] Answer: \answerYes{} 
    \item[] Justification: The paper reports that TabPFN finetuning and Llama-3-8B QLoRA finetuning were performed on a single NVIDIA A100 GPU, and Appendix~\ref{app:tabpfn} reports per-horizon TabPFN finetuning times for the released prototype-undersampling checkpoints. Appendix~\ref{app:llama} provides the Llama-3-8B training configuration, including 4-bit loading, batch size, gradient accumulation, number of epochs, and train/validation sample counts. Classical baselines are comparatively lightweight tabular models; their hyperparameter grids are reported in Appendix~\ref{app:baselines}.
    \item[] Guidelines:
    \begin{itemize}
        \item The answer \answerNA{} means that the paper does not include experiments.
        \item The paper should indicate the type of compute workers CPU or GPU, internal cluster, or cloud provider, including relevant memory and storage.
        \item The paper should provide the amount of compute required for each of the individual experimental runs as well as estimate the total compute. 
        \item The paper should disclose whether the full research project required more compute than the experiments reported in the paper (e.g., preliminary or failed experiments that didn't make it into the paper). 
    \end{itemize}
    
\item {\bf Code of ethics}
    \item[] Question: Does the research conducted in the paper conform, in every respect, with the NeurIPS Code of Ethics \url{https://neurips.cc/public/EthicsGuidelines}?
    \item[] Answer: \answerYes{} 
    \item[] Justification: The research uses corporate financial records and evaluates bankruptcy/distress prediction methods for benchmarking purposes. The paper does not involve human-subject experiments, deception, harmful data collection, or deployment of an automated decision system, and the released benchmark is intended to support reproducible evaluation.
    \item[] Guidelines:
    \begin{itemize}
        \item The answer \answerNA{} means that the authors have not reviewed the NeurIPS Code of Ethics.
        \item If the authors answer \answerNo, they should explain the special circumstances that require a deviation from the Code of Ethics.
        \item The authors should make sure to preserve anonymity (e.g., if there is a special consideration due to laws or regulations in their jurisdiction).
    \end{itemize}

\item {\bf Broader impacts}
    \item[] Question: Does the paper discuss both potential positive societal impacts and negative societal impacts of the work performed?
    \item[] Answer: \answerYes{} 
    \item[] Justification: The paper discusses responsible use in Section~\ref{sec:limitations}. It identifies positive uses of V4FinBench for reproducible benchmarking, financial-risk monitoring, and method development, while also noting risks from direct automated decision-making, including possible effects on credit access, supplier relationships, investment decisions, and regulatory attention. The paper states that deployment would require additional validation, fairness analysis, monitoring, and human oversight.
    \item[] Guidelines:
    \begin{itemize}
        \item The answer \answerNA{} means that there is no societal impact of the work performed.
        \item If the authors answer \answerNA{} or \answerNo, they should explain why their work has no societal impact or why the paper does not address societal impact.
        \item Examples of negative societal impacts include potential malicious or unintended uses (e.g., disinformation, generating fake profiles, surveillance), fairness considerations (e.g., deployment of technologies that could make decisions that unfairly impact specific groups), privacy considerations, and security considerations.
        \item The conference expects that many papers will be foundational research and not tied to particular applications, let alone deployments. However, if there is a direct path to any negative applications, the authors should point it out. For example, it is legitimate to point out that an improvement in the quality of generative models could be used to generate Deepfakes for disinformation. On the other hand, it is not needed to point out that a generic algorithm for optimizing neural networks could enable people to train models that generate Deepfakes faster.
        \item The authors should consider possible harms that could arise when the technology is being used as intended and functioning correctly, harms that could arise when the technology is being used as intended but gives incorrect results, and harms following from (intentional or unintentional) misuse of the technology.
        \item If there are negative societal impacts, the authors could also discuss possible mitigation strategies (e.g., gated release of models, providing defenses in addition to attacks, mechanisms for monitoring misuse, mechanisms to monitor how a system learns from feedback over time, improving the efficiency and accessibility of ML).
    \end{itemize}
    
\item {\bf Safeguards}
    \item[] Question: Does the paper describe safeguards that have been put in place for responsible release of data or models that have a high risk for misuse (e.g., pre-trained language models, image generators, or scraped datasets)?
    \item[] Answer: \answerNA{} 
    \item[] Justification: The paper releases a structured corporate financial benchmark and TabPFN checkpoints for bankruptcy/distress prediction, not a high-risk generative model or scraped dataset containing unsafe media or personal content. The main risks are downstream misuse or overreliance in financial decision-making, which are better addressed under broader impacts and limitations rather than through release safeguards.
    \item[] Guidelines:
    \begin{itemize}
        \item The answer \answerNA{} means that the paper poses no such risks.
        \item Released models that have a high risk for misuse or dual-use should be released with necessary safeguards to allow for controlled use of the model, for example by requiring that users adhere to usage guidelines or restrictions to access the model or implementing safety filters. 
        \item Datasets that have been scraped from the Internet could pose safety risks. The authors should describe how they avoided releasing unsafe images.
        \item We recognize that providing effective safeguards is challenging, and many papers do not require this, but we encourage authors to take this into account and make a best faith effort.
    \end{itemize}

\item {\bf Licenses for existing assets}
    \item[] Question: Are the creators or original owners of assets (e.g., code, data, models), used in the paper, properly credited and are the license and terms of use explicitly mentioned and properly respected?
    \item[] Answer: \answerYes{} 
    \item[] Justification: All existing assets used in the paper are credited via citations and listed with their licenses and source URLs in Appendix~\ref{app:licenses}. This covers the EMIS data source underlying V4FinBench (acquired under an individual data-purchase agreement, with explicit permission from EMIS for the public release of the derived benchmark), the American Bankruptcy Dataset~\cite{lombardo2022machine} used in the transfer experiment (CC0~1.0), the Meta-Llama-3-8B model~\cite{llama3_modelcard} (Meta Llama~3 Community License), TabPFN v2~\cite{hollmann2025tabpfn}, and the gradient-boosting libraries XGBoost~\cite{chen2016xgboost} (Apache~2.0), CatBoost~\cite{prokhorenkova2018catboost} (Apache~2.0), and LightGBM~\cite{ke2017lightgbm} (MIT). All assets are used in accordance with their respective licenses.
    \item[] Guidelines:
    \begin{itemize}
        \item The answer \answerNA{} means that the paper does not use existing assets.
        \item The authors should cite the original paper that produced the code package or dataset.
        \item The authors should state which version of the asset is used and, if possible, include a URL.
        \item The name of the license (e.g., CC-BY 4.0) should be included for each asset.
        \item For scraped data from a particular source (e.g., website), the copyright and terms of service of that source should be provided.
        \item If assets are released, the license, copyright information, and terms of use in the package should be provided. For popular datasets, \url{paperswithcode.com/datasets} has curated licenses for some datasets. Their licensing guide can help determine the license of a dataset.
        \item For existing datasets that are re-packaged, both the original license and the license of the derived asset (if it has changed) should be provided.
        \item If this information is not available online, the authors are encouraged to reach out to the asset's creators.
    \end{itemize}

\item {\bf New assets}
    \item[] Question: Are new assets introduced in the paper well documented and is the documentation provided alongside the assets?
    \item[] Answer: \answerYes{} 
    \item[] Justification: The paper introduces V4FinBench and documents its data source, coverage, distress definition, multi-horizon construction, evaluation protocol, feature set, and limitations in Sections~\ref{sec:dataset} and~\ref{sec:limitations} and Appendix~\ref{app:data-source}. The released Kaggle dataset, GitHub label-derivation code, fold indices, and Hugging Face TabPFN checkpoints provide supporting documentation and artifacts alongside the benchmark.
    \item[] Guidelines:
    \begin{itemize}
        \item The answer \answerNA{} means that the paper does not release new assets.
        \item Researchers should communicate the details of the dataset\slash code\slash model as part of their submissions via structured templates. This includes details about training, license, limitations, etc. 
        \item The paper should discuss whether and how consent was obtained from people whose asset is used.
        \item At submission time, remember to anonymize your assets (if applicable). You can either create an anonymized URL or include an anonymized zip file.
    \end{itemize}

\item {\bf Crowdsourcing and research with human subjects}
    \item[] Question: For crowdsourcing experiments and research with human subjects, does the paper include the full text of instructions given to participants and screenshots, if applicable, as well as details about compensation (if any)? 
    \item[] Answer: \answerNA{} 
    \item[] Justification: The paper does not involve crowdsourcing, user studies, surveys, annotation tasks, or experiments with human subjects. The benchmark is constructed from corporate financial records, not collected from study participants.
    \item[] Guidelines:
    \begin{itemize}
        \item The answer \answerNA{} means that the paper does not involve crowdsourcing nor research with human subjects.
        \item Including this information in the supplemental material is fine, but if the main contribution of the paper involves human subjects, then as much detail as possible should be included in the main paper. 
        \item According to the NeurIPS Code of Ethics, workers involved in data collection, curation, or other labor should be paid at least the minimum wage in the country of the data collector. 
    \end{itemize}

\item {\bf Institutional review board (IRB) approvals or equivalent for research with human subjects}
    \item[] Question: Does the paper describe potential risks incurred by study participants, whether such risks were disclosed to the subjects, and whether Institutional Review Board (IRB) approvals (or an equivalent approval/review based on the requirements of your country or institution) were obtained?
    \item[] Answer: \answerNA{} 
    \item[] Justification: The paper does not involve crowdsourcing or research with human subjects, so there are no study participants, participant-facing risks, consent procedures, or IRB approval requirements to report.
    \item[] Guidelines:
    \begin{itemize}
        \item The answer \answerNA{} means that the paper does not involve crowdsourcing nor research with human subjects.
        \item Depending on the country in which research is conducted, IRB approval (or equivalent) may be required for any human subjects research. If you obtained IRB approval, you should clearly state this in the paper. 
        \item We recognize that the procedures for this may vary significantly between institutions and locations, and we expect authors to adhere to the NeurIPS Code of Ethics and the guidelines for their institution. 
        \item For initial submissions, do not include any information that would break anonymity (if applicable), such as the institution conducting the review.
    \end{itemize}

\item {\bf Declaration of LLM usage}
    \item[] Question: Does the paper describe the usage of LLMs if it is an important, original, or non-standard component of the core methods in this research? Note that if the LLM is used only for writing, editing, or formatting purposes and does \emph{not} impact the core methodology, scientific rigor, or originality of the research, declaration is not required.
    \item[] Answer: \answerYes{} 
    \item[] Justification: Llama-3-8B is one of the evaluated methods in the paper, and its use is described in Section~\ref{sec:llama}, Section~\ref{sec:exp-llama}, and Appendix~\ref{app:llama}. These sections specify the model, QLoRA finetuning setup, serialization format, training configuration, and evaluation procedure.
    \item[] Guidelines:
    \begin{itemize}
        \item The answer \answerNA{} means that the core method development in this research does not involve LLMs as any important, original, or non-standard components.
        \item Please refer to our LLM policy in the NeurIPS handbook for what should or should not be described.
    \end{itemize}

\end{enumerate}

\end{document}